\documentclass[10pt,journal,compsoc]{IEEEtran}
\usepackage{amsmath,amsfonts}
\usepackage{algorithmic}
\usepackage{algorithm}
\usepackage{array}
\usepackage[caption=false,font=normalsize,labelfont=sf,textfont=sf]{subfig}
\usepackage{textcomp}
\usepackage{stfloats}
\usepackage[hidelinks,colorlinks]{hyperref} 
\usepackage{verbatim}
\usepackage{graphicx}
\usepackage{xspace,xpunctuate}
\usepackage{cite}
\usepackage{xcolor,colortbl}
\usepackage{booktabs}
\usepackage{orcidlink}
\usepackage{cleveref}
\usepackage{multirow}
\usepackage{pifont}
\usepackage{comment}
\usepackage{scalerel}
\usepackage{fontawesome5} 
\usepackage{twemojis} 
\usepackage{capt-of}
\usepackage{pdflscape}
\usepackage{pdfpages} 
\newcommand{\xmark}{\ding{55}}
\newcommand{\cmark}{\ding{51}}
\newcommand{\rl}{{\texttwemoji{brain}}\xspace}                     
\newcommand{\dis}{{\texttwemoji{1f4ca}}\xspace}                    
\newcommand{\yay}{{\color{green}\cmark}}
\newcommand{\nej}{{\color{red}\xmark}}
\newcommand{\tba}{{\color{blue}TBA}}
\newcommand{\non}{}

\def\oaZ{\scalerel{\includegraphics[decodearray={1 0.796 1 0.89 1 1}]{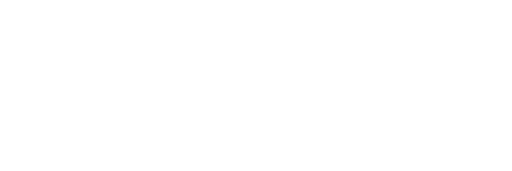}}{\rule[-2\LMpt]{0pt}{8\LMpt}}}
\def\oaO{\scalerel{\includegraphics[decodearray={1 0.796 1 0.89 1 1}]{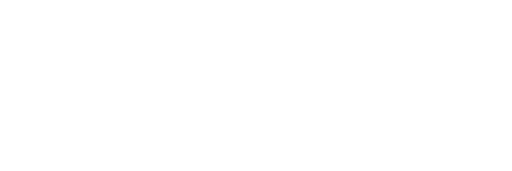}}{\rule[-2\LMpt]{0pt}{8\LMpt}}}
\def\oaT{\scalerel{\includegraphics[decodearray={1 0.796 1 0.89 1 1}]{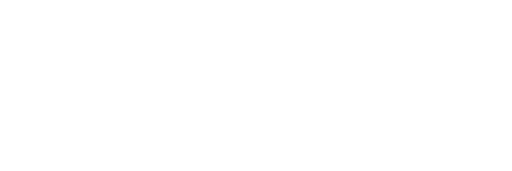}}{\rule[-2\LMpt]{0pt}{8\LMpt}}}
\def\oaH{\scalerel{\includegraphics[decodearray={1 0.796 1 0.89 1 1}]{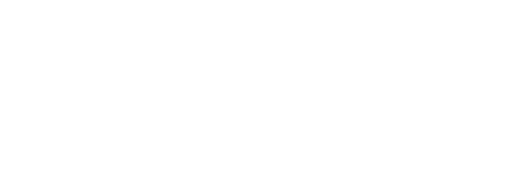}}{\rule[-2\LMpt]{0pt}{8\LMpt}}}

\def\srZ{\scalerel{\includegraphics[decodearray={1 1 1 0.831 1 0.808}]{figures/icon_0.png}}{\rule[-2\LMpt]{0pt}{8\LMpt}}}
\def\srO{\scalerel{\includegraphics[decodearray={1 1 1 0.831 1 0.808}]{figures/icon_1.png}}{\rule[-2\LMpt]{0pt}{8\LMpt}}}
\def\srT{\scalerel{\includegraphics[decodearray={1 1 1 0.831 1 0.808}]{figures/icon_2.png}}{\rule[-2\LMpt]{0pt}{8\LMpt}}}
\def\srH{\scalerel{\includegraphics[decodearray={1 1 1 0.831 1 0.808}]{figures/icon_3.png}}{\rule[-2\LMpt]{0pt}{8\LMpt}}}

\def\nsZ{\scalerel{\includegraphics[decodearray={1 1 1 0.831 1 0.808}]{figures/icon_0.png}}{\rule[-2\LMpt]{0pt}{8\LMpt}}}
\def\nsO{\scalerel{\includegraphics[decodearray={1 1 1 0.831 1 0.808}]{figures/icon_1.png}}{\rule[-2\LMpt]{0pt}{8\LMpt}}}
\def\nsT{\scalerel{\includegraphics[decodearray={1 1 1 0.831 1 0.808}]{figures/icon_2.png}}{\rule[-2\LMpt]{0pt}{8\LMpt}}}
\def\nsH{\scalerel{\includegraphics[decodearray={1 1 1 0.831 1 0.808}]{figures/icon_3.png}}{\rule[-2\LMpt]{0pt}{8\LMpt}}}

\def\abZ{\scalerel{\includegraphics[decodearray={1 0.98 1 0.906 1 0.725}]{figures/icon_0.png}}{\rule[-2\LMpt]{0pt}{8\LMpt}}}
\def\abO{\scalerel{\includegraphics[decodearray={1 0.98 1 0.906 1 0.725}]{figures/icon_1.png}}{\rule[-2\LMpt]{0pt}{8\LMpt}}}
\def\abT{\scalerel{\includegraphics[decodearray={1 0.98 1 0.906 1 0.725}]{figures/icon_2.png}}{\rule[-2\LMpt]{0pt}{8\LMpt}}}
\def\abH{\scalerel{\includegraphics[decodearray={1 0.98 1 0.906 1 0.725}]{figures/icon_3.png}}{\rule[-2\LMpt]{0pt}{8\LMpt}}}

\newsavebox\texticonbox
\newsavebox\imgiconbox
\newsavebox\fineiconbox

\savebox\texticonbox{\includegraphics{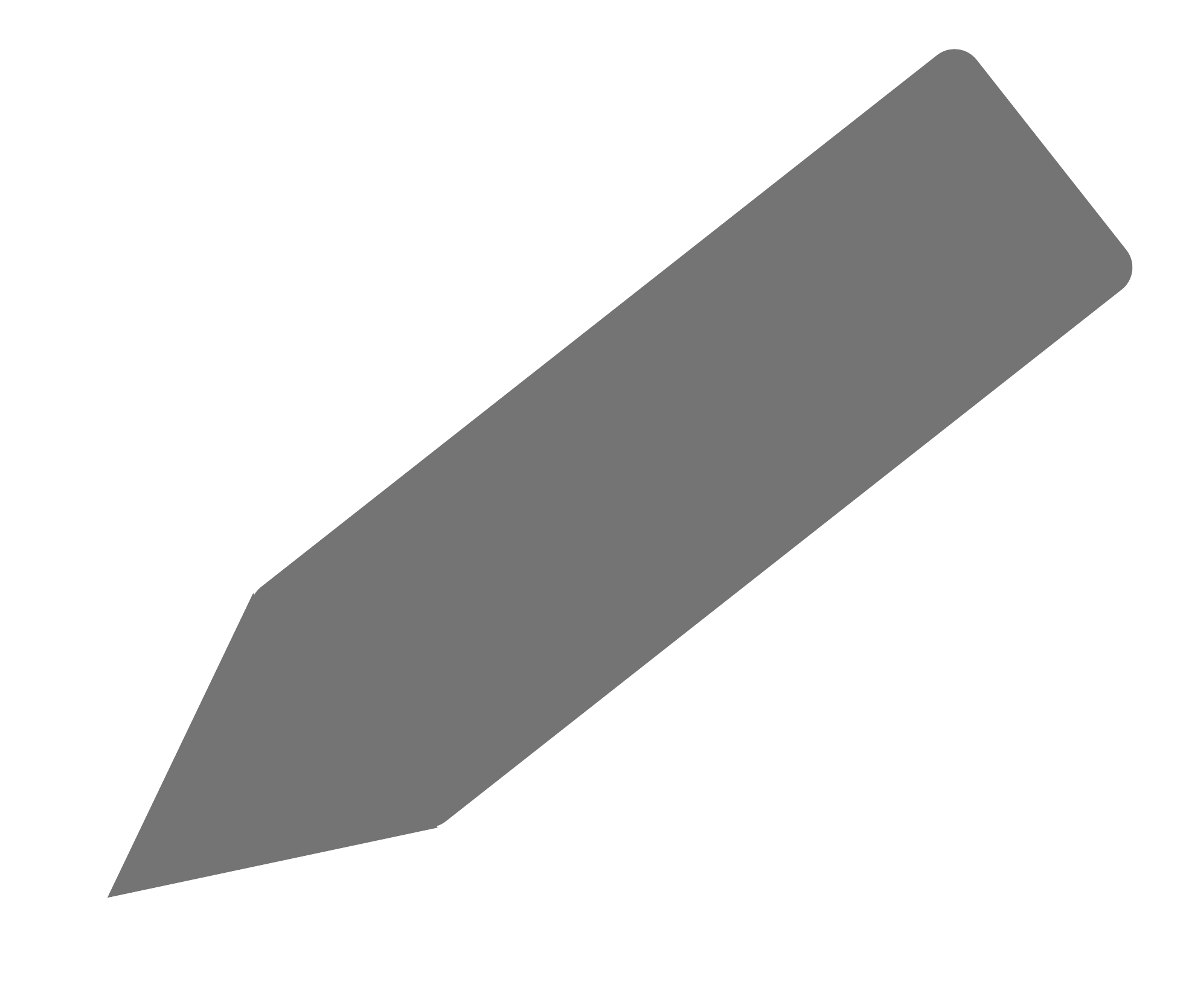}}
\savebox\imgiconbox{\includegraphics{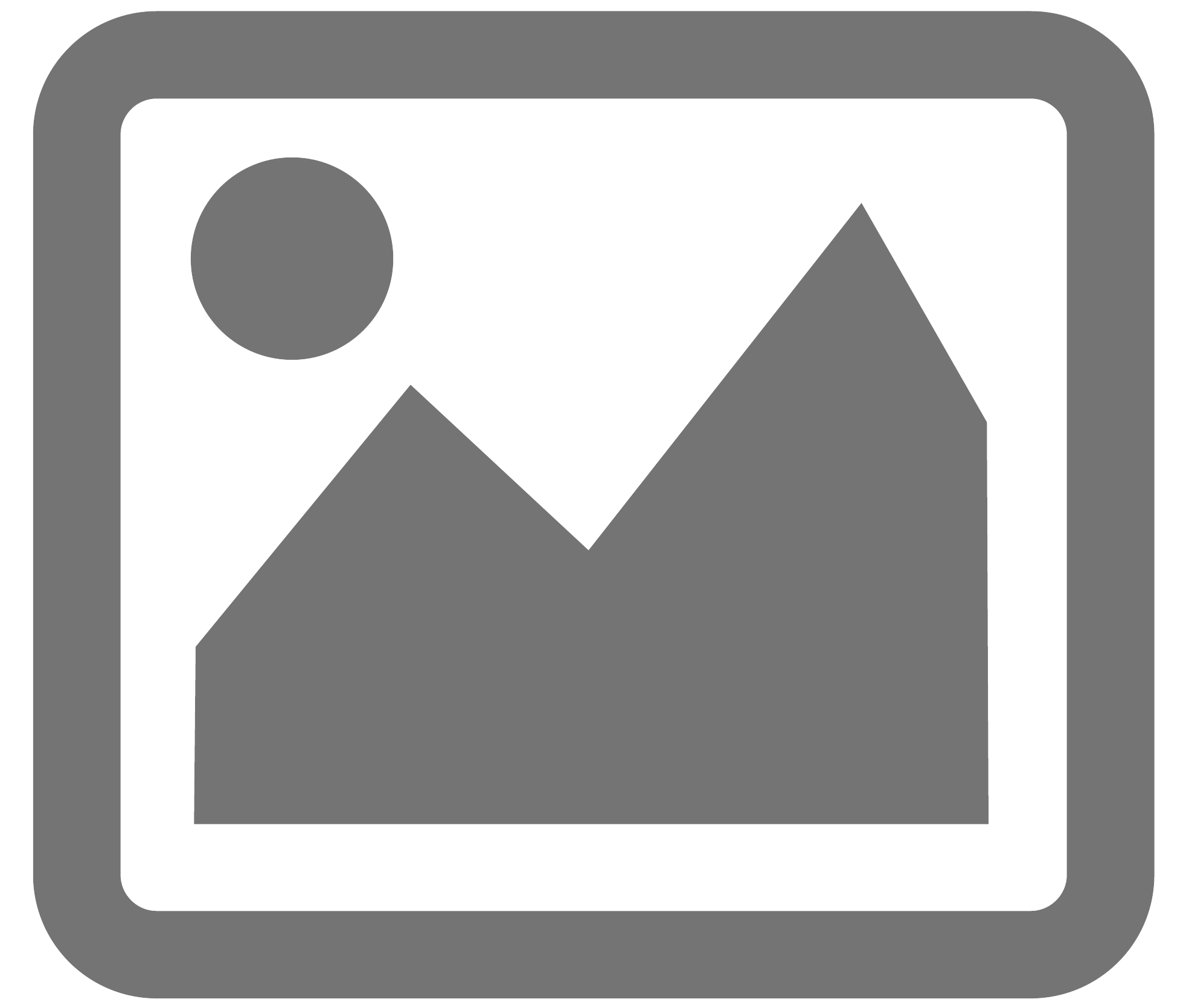}}
\savebox\fineiconbox{\includegraphics{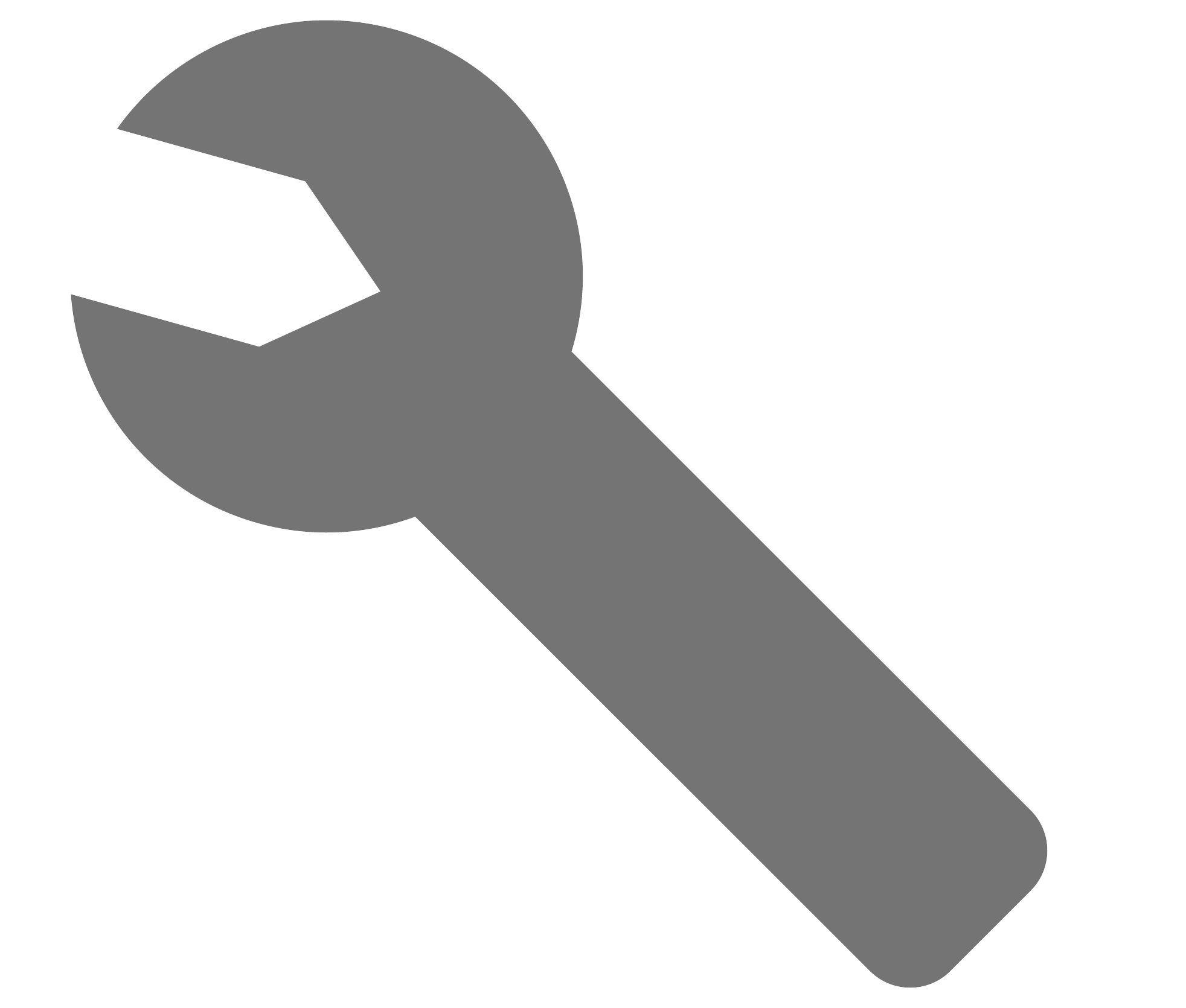}}

\def\txt{\scalerel{\usebox{\texticonbox}}{\rule[-2\LMpt]{0pt}{8\LMpt}}}
\def\img{\scalerel{\usebox{\imgiconbox}}{\rule[-2\LMpt]{0pt}{8\LMpt}}}
\def\ft{\scalerel{\usebox{\fineiconbox}}{\rule[-2\LMpt]{0pt}{8\LMpt}}}

\definecolor{tticolor}{HTML}{DEE6EF}
\definecolor{ittcolor}{HTML}{C7D6E5}
\definecolor{embcolor}{HTML}{7498BE}
\definecolor{concolor}{HTML}{A2BAD4}
\definecolor{discolor}{HTML}{EB9293}
\definecolor{sincolor}{HTML}{F1AFB0}
\definecolor{graycolor}{gray}{0.9}
\definecolor{oacolor}{HTML}{B2D6FF}
\definecolor{srcolor}{HTML}{FFD4CE}
\definecolor{nscolor}{HTML}{FFD4CE}
\definecolor{abcolor}{HTML}{FAE7B9}

\newcommand{\etal}{ et al.~}

\newcommand{\ie}{{i.e.,}\xspace}
\newcommand{\cf}{{cf.}\xspace}
\newcommand{\eg}{{e.g.,}\xspace}

\newcommand{\etc}{{etc\xperiod}\xspace}

\begin{document}

\title{A Survey on Quality Metrics\\ for Text-to-Image Generation}
\author{Sebastian Hartwig\orcidlink{0000-0001-8642-2789}, 
        Dominik Engel\orcidlink{0000-0002-5766-7215},
        Leon Sick\orcidlink{0009-0004-6524-0715},
        Hannah Kniesel\orcidlink{0000-0001-5898-8152},
        Tristan Payer\orcidlink{0009-0005-9602-3366}, \\
        Poonam Poonam\orcidlink{0009-0002-0472-229X},
        Michael Glöckler\orcidlink{0000-0002-4436-918X},
        Alex Bäuerle\orcidlink{0000-0003-3886-8799},
        Timo Ropinski\orcidlink{0000-0002-7857-5512}
\IEEEcompsocitemizethanks{\IEEEcompsocthanksitem S. Hartwig, D. Engel, L. Sick, H. Kniesel, T. Payer, P. Poonam, M. Glöckler, T. Ropinski
are with Visual Computing Group located at Ulm University\protect\\
E-mail: \{forename\}.\{surname\}@uni-ulm.de
\IEEEcompsocthanksitem Alex Bäuerle is Postdoctoral Researcher at Carnegie Mellon University
E-mail: alex@a13x.io
\IEEEcompsocthanksitem Website: \url{https://huggingface.co/spaces/kopetri/text-to-image-evaluation}
}
\thanks{Manuscript received January 28, 2025.}}

%
%

\markboth{preprint}%
{Hartwig \MakeLowercase{\textit{et al.}}: Visual Computing Group located at Ulm University}


%



\IEEEteaser{
    \centering
    \includegraphics[width=0.9\linewidth]{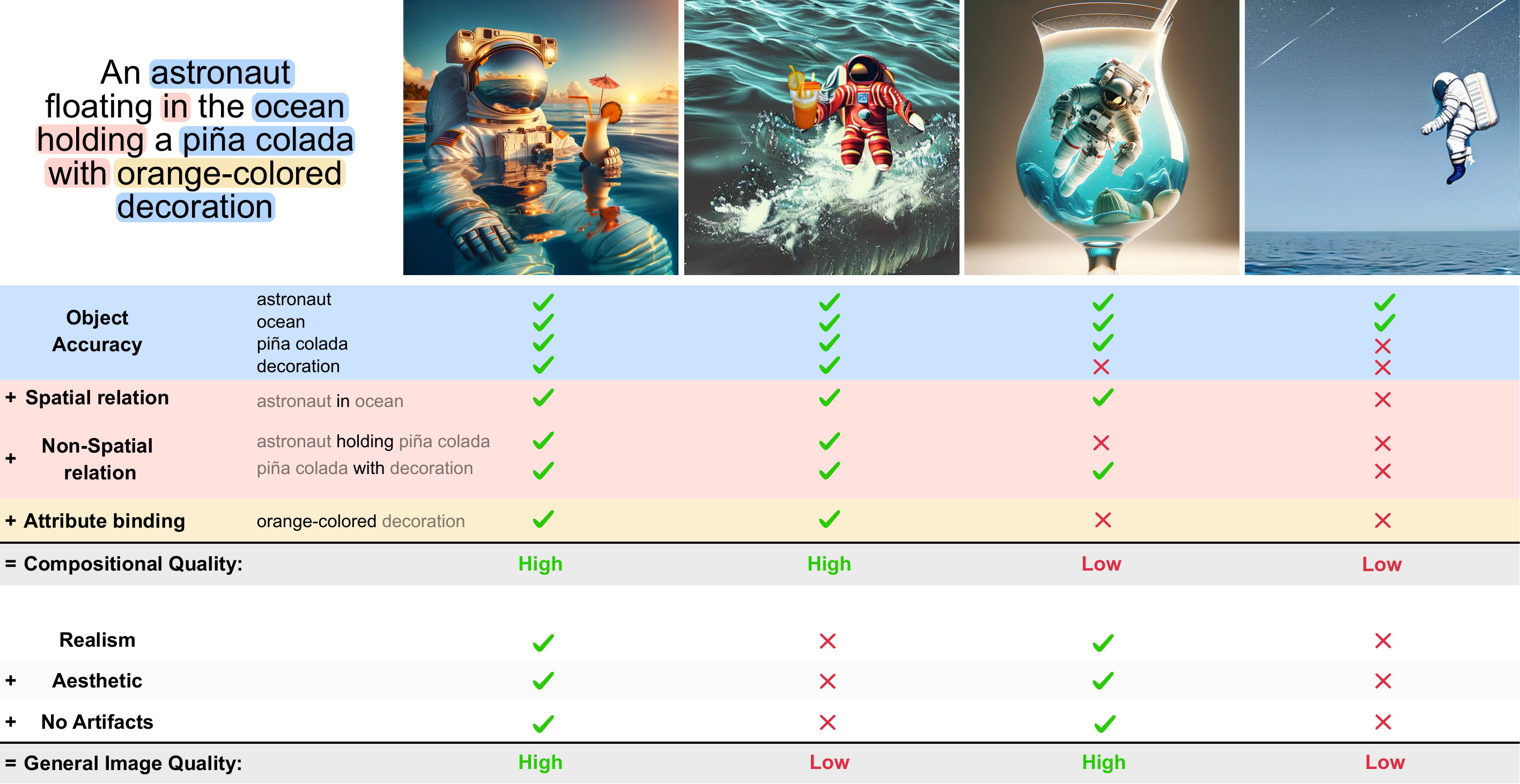}
    \setcounter{figure}{0}
    \vspace*{-1mm}
    \captionof{figure}{Evaluating AI-based text-to-image generation requires two types of quality measures that contribute to an overall image quality score. \textit{Compositional Quality} measures how well the image reflects the composition defined in the text prompt, while \textit{General Image Quality} measures the overall quality of the image. For both types, several different aspects have to be considered depending on the desired use.\label{fig:teaser}}
}

\IEEEtitleabstractindextext{%
\begin{abstract}
AI-based text-to-image models do not only excel at generating realistic images, they also give designers more and more fine-grained control over the image content. Consequently, these approaches have gathered increased attention within the computer graphics research community, which has been historically devoted towards traditional rendering techniques, that offer precise control over scene parameters (e.g., objects, materials, and lighting). While the quality of conventionally rendered images is assessed through well established image quality metrics, such as SSIM or PSNR, the unique challenges of text-to-image generation require other, dedicated quality metrics. These metrics must be able to not only measure overall image quality, but also how well images reflect given text prompts, whereby the control of scene and rendering parameters is interweaved. Within this survey, we provide a comprehensive overview of such text-to-image quality metrics, and propose a taxonomy to categorize these metrics. Our taxonomy is grounded in the assumption, that there are two main quality criteria, namely compositional quality and general quality, that contribute to the overall image quality. Besides the metrics, this survey covers dedicated text-to-image benchmark datasets, over which the metrics are frequently computed. Finally, we identify limitations and open challenges in the field of text-to-image generation, and derive guidelines for practitioners conducting text-to-image evaluation.
\end{abstract}

}

\maketitle

\IEEEdisplaynontitleabstractindextext

%
\IEEEpeerreviewmaketitle

\IEEEraisesectionheading{\section{Introduction}\label{sec:introduction}}

\IEEEPARstart{T}{he} rapidly evolving landscape of AI-based text-to-image (T2I) generation models has emerged as a pivotal area in computer graphics~\cite{10.1145/3528223.3530104,Feng_2023,xu2024dreamanime,xing2024make}, computer vision and natural language processing~\cite{10.1145/3461353.3461388,zhang2023texttoimage,10081412}. For the computer graphics community, T2I generation opens new avenues for developing more intuitive and user-friendly interfaces for graphics software, enabling artists and designers to generate virtual imagery simply through textual descriptions. Furthermore, it pushes the boundaries of traditional rendering techniques by integrating linguistic context into visual content~\cite{po2023state,Zeng2024IntentTunerAI}, which can revolutionize how visual effects are created, how narratives are visualized, and how interactive media is produced. After significantly lowering the hardware requirements for T2I models through latent diffusion, an increasing number of publicly available T2I models, such as DALL-E, ImageFX, DreamStudio, Midjourney, as well as Flux, became accessible not only to researchers, but also to novices. Thus, as the technology progresses, it starts to diminish the time and technical barriers traditionally involved in high-quality graphics production.

As the demand for seamless integration between textual and visual information intensifies, understanding the intricate mechanisms influencing T2I generation becomes imperative. To do so, T2I quality metrics are an essential tool, as they allow for objectively evaluating T2I generation models. Unfortunately, defining requirements for T2I image quality metrics is not straightforward. Realism is undoubtedly the major aspect targeted by researchers~\cite{8195348}. However, the interpretation of realism highly depends on the text conditioning, \eg{} an image can be photorealistic, realistic in the context of a manga or realistic in the style of Pablo Picasso. Other aspects that contribute to high-quality images include aesthetics~\cite{schuhmann2022laion}, human preferences~\cite{xu2023imagereward,Kirstain2023PickaPicAO,Wu2023HumanPS,hartwig2022learning}, naturalism, and the principles of photography, such as balance, harmony, closure, movement, color, pattern, contrast, negative space, and grouping. Although some of these aspects may be quantitatively measurable, many are abstract, complex, and therefore difficult to measure. However, natural language can depict these aspects in great detail, and there are many talented authors who generate creative descriptions of sceneries. Hence, detecting and measuring the quality of these abstract yet well-described aspects presents a challenge to researchers in the field of text-conditioned image generation.

Within this survey, we aim to review and categorize T2I quality metrics comprehensively with the goal to provide both an overarching perspective and actionable insights to assist researchers and practitioners in evaluating T2I generation models effectively. To help structure the existing literature, we define the overall quality of an image conditioned on a text prompt as a combination of \textit{general quality} and \textit{compositional quality}, where the latter measures the degree of alignment between the text and the image (see Figure~\ref{fig:teaser}). A high compositional quality score can only be achieved if all details described in a text prompt are visually represented in the image, while a high general quality does not have any implications on how closely the image content reflects the text prompt. By considering these two main image quality contributors, we are able to review existing T2I quality metrics in a structured manner, and to derive a taxonomy classifying existing T2I quality metrics. While the taxonomy has emerged from the reviewed quality metrics, we will present it first within Section~\ref{sec:taxonomy}, as we believe that it is an important tool to understand the field of T2I quality metrics, and ultimately guide the reader through this survey. After the taxonomy has been laid out, the reviewed T2I quality metrics are presented in Section~\ref{sec:metrics}, which is structured according to the main categories of our taxonomy. To allow for an objective comparison of T2I generation models, not only are the used quality metrics of importance, but also the datasets on which these metrics are evaluated. Therefore, we will cover T2I evaluation data sets within Section~\ref{sec:datasets}. Based on the reviewed metrics and datasets, we will further outline open challenges related to the evaluation of T2I models (see Section~\ref{sec:challenges}), and provide guidelines for practitioners and researchers evaluating T2I models (see Section~\ref{sec:guidelines}). In Section 1 of our supplementary material, we discuss methods that apply T2I quality metrics to optimize image generation. In Section 2, we provide experimental results from an investigation of a selection of six human preference metrics. Finally, the survey will conclude in Section~\ref{sec:conclusion}.

 

\section{Taxonomy}\label{sec:taxonomy}
\begin{figure*}[ht]
    \centering
    \includegraphics[width=0.85\textwidth]{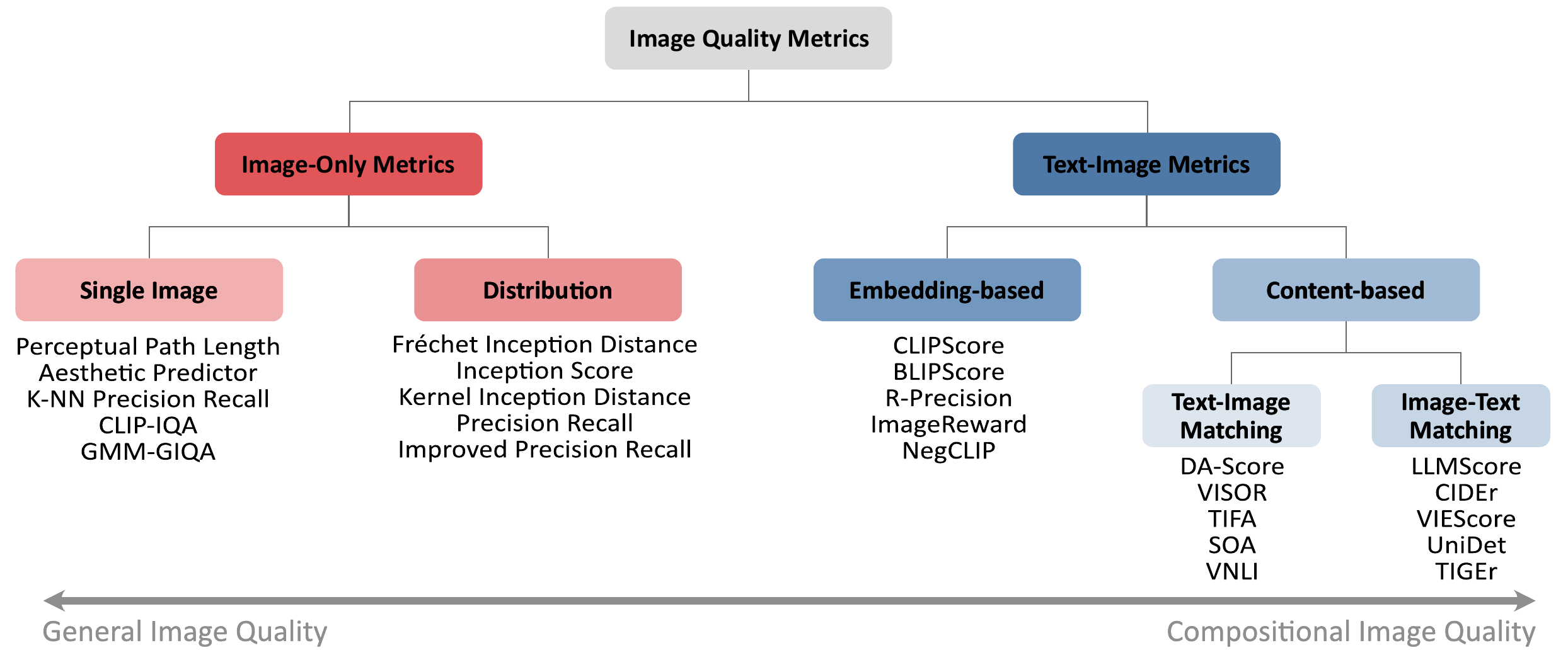}
    \caption{Proposed taxonomy and examples for T2I evaluation metrics. Two categories of metrics need to be distinguished: image-only and text-conditioned image quality metrics. Image quality metrics can measure two different qualities which correlate with this categorization, namely general image quality and compositional quality.}
    \label{fig:taxonomy}
\end{figure*}

Based on the reviewed T2I image quality metrics (see Section~\ref{sec:metrics}), we have derived a taxonomy, which helps readers to gain an overview of the field of these metrics. Within this taxonomy, we on the one hand consider \textit{image-only quality metrics}, that are designed to measure image quality by solely considering images and thus not consider text prompts (red boxes in \Cref{fig:taxonomy}). In contrast, \textit{text-image quality metrics} consider the agreement between image content and text prompt (blue boxes in \Cref{fig:taxonomy}). While image-only quality metrics can solely be used to express \text{general image quality}, text-image quality metrics can be used to measure \textit{compositional image quality}, which can be considered of a finer granular nature, as individual entities are taken into account. Since both, general image quality and compositional image quality, are relevant to assess a T2I generation model, often both qualities are assessed, and an average score of general and compositional image quality is reported. As high compositional quality does not automatically result in high general image quality and vice versa, both scores should though also be reported alongside their average.

Image-only quality metrics and text-image quality metrics naturally differ based on their input. Image-only quality metrics solely process the image $x$ as input, while text-image quality metrics process the image $x$ together with the text prompt $t$ as input. Besides this main distinction, other metric properties are relevant during categorization. Therefore, when reviewing the discussed metric papers, we have performed a coding of the most essential components of each metric, and used these components to further inform our taxonomy. Accordingly, the presented taxonomy categorizes T2I quality metrics based on their operating data structure (embeddings vs. contents), measured aspects (general quality vs. compositional quality), scope (distribution of images vs. single images), and conditions (image-only vs. text-image) used. In the following, we will describe the categories of our taxonomy with respect to these components. To enable the reader to better relate the described metrics to each other, we further introduce a mathematical notation to formalize their description. The variables used in this notation are outlined in Table~\ref{tab:variables}. 

\begin{table}[b]
  \centering
  \begin{tabular}{cl}
    \toprule
    Variable     & Description \\
    \midrule
    $t$          & text prompt \\
    $x$          & Image \\
    $X$          & Set of images \\
    $f_t$        & Text embedding vector \\
    $f_i$        & Image embedding vector \\
    $D$          & Distance measure (usually non-trainable) \\
    $F^{\theta}$ & Neural model with trainable parameters $\theta$ \\
    $\Phi_T$     & Text dissection process \\
    $\Phi_I$     & Image dissection process \\
    $S_T$        & Separated content elements of a prompt \\
    $S_I$        & Separated content elements of an image \\
    $\eta$       & Image captioning model \\
    \bottomrule
  \end{tabular}
  \vspace{3pt}
  \caption{Description of variables used for our mathematical notation describing image and text-image quality metrics.}
  \label{tab:variables}
\end{table}

\subsection{Image-Only Quality Metrics}\label{ssec:tax-image-based}
Pure \textit{image-only quality metrics} do not take into account the textual condition $t$ when evaluating generated images, and can therefore only evaluate \textit{general image quality}, as opposed to \textit{compositional image quality}. We define \textit{general image quality} as quantifying a certain aspect globally for a single image, \eg{} realism, aesthetics, and human preferences, for which ground truth can be collected by asking human raters for their judgments. Usually, this is done by conducting a large-scale crowd-sourced study where images are ranked by a large group of human observers, which is a high-effort endeavor. Hence, practitioners use the acquired human annotations to develop deep learning-based evaluation models that are designed to imitate such human judgments, \eg{} Aesthetic Predictor~\cite{schuhmann2022laion}, PAL4VAST~\cite{zhang2023perceptual,zhang2022perceptual}, and Human Viewpoint Preferences~\cite{hartwig2022learning}. In many scenarios, these image quality metrics play an important role, as image quality is often assessed independently from how well the image content depicts a given text prompt. For instance, an image that appears photorealistic but disregards the textual content may receive a high general quality score but a low compositional quality score. Conversely, an image that accurately represents all objects and relationships described in the prompt might still look artificial, leading to a low general quality score but a high compositional quality score. In \Cref{fig:teaser}, we illustrate such examples.

Among the image quality metrics, techniques can be further divided based on how the image information is considered. Accordingly, we further divide the image quality metrics category into two subcategories, one processing single images only, and one processing distributions. We refer to these subcategories as \textit{single image quality metrics} and \textit{distribution metrics}, and will discuss them in more detail below.

\subsubsection{Single Image Quality Metrics}
\textit{Single image quality metrics} measure quality for individual images by analyzing an image $x$ based on its structural and semantic composition. They are also referred to as no-reference image quality metrics, as no ground truth or other reference image is used to compare the image to. Instead, single image quality metrics $QM_{single}$ extract features from the image and subsequently infer quality, which is measured by a quality measure $D_I$:

\begin{equation}
    QM_{single}(x) = D_I(x)
\end{equation}
Recent approaches, which can be categorized as single image quality metrics, often rely on a fine-tuned image model $F_s^{\theta}$ that is trained to predict human judgments, \eg{} LAION Aesthetic Predictor~\cite{schuhmann2022laion}, perceptual artifact localization (PAL)~\cite{zhang2022perceptual,zhang2023perceptual} or human viewpoint preferences (HVP)~\cite{hartwig2022learning}, with $\theta$ being the fine-tuned weights. Thus, the metric $QM_{single}(x)$ becomes dependent on some learned parameters $\theta$, and can be formulated as:
\begin{equation}
    QM^{\theta}_{single}(x) = F_s^{\theta}(x)
\end{equation}

\subsubsection{Distribution Metrics}
\textit{Distribution metrics} are focused on evaluating a T2I generation model, rather than individual outputs. Thus, while the T2I model is treated as a black box, its quality is evaluated based on the output sample distribution $p(X_g)$. During evaluation, the differences between the distribution $p(X_g)$ of generated data $X_g$ and the distribution $p(X_t)$ of given, usually real-world, target data $X_t$ are analyzed. A general definition of a distribution-based quality metric $QM_{distribution}$ can be formulated as follows:
\begin{equation}
    QM_{distribution}(X_g, X_t) = D_D(p(X_g), p(X_t)) 
    \label{eq:distribution_metric}
\end{equation}
where $D_D$ is a statistical distance or divergence measure between two probability distributions.

Performing direct calculations $D_D$ in the high-dimensional image space is often intractable. Therefore, images are mapped to a lower-dimensional feature space $f_g = F^{\theta}(X_g)$ and $f_t = F^{\theta}(X_t)$ using a pre-trained feature extractor $F^{\theta}$ with trainable parameters $\theta$, such as an intermediate layer of a convolutional neural network. Thus, when considering this projection, $QM^{\theta}_{distribution}$ can be formalized as follows:
\begin{equation}
    QM^{\theta}_{distribution}(X_g, X_t) = D_D(p(F^{\theta}(X_g)), p(F^{\theta}(X_t)))
\end{equation}

\noindent Frequently used distribution metrics are the Inception Score (IS)~\cite{NIPS2016_8a3363ab} and the Fréchet Inception Distance (FID)~\cite{heusel2017gans}.

\subsection{Text-Image Quality Metrics}\label{ssec:t2i-metrics}
\textit{Text-image quality metrics} measure the degree to which an image depicts a textual prompt that has been used for its generation. Thus, in contrast to image-only quality metrics, they can measure \textit{compositional quality} by dissecting the prompt and the image into multiple text-image pairs and evaluating the matching of these pairs. In particular, quality is measured by analyzing the alignment between prompt specification and content depicted in the image, \eg{} through measures such as \textbf{object accuracy (OA)}, \textbf{spatial relation (S)}, \textbf{non-spatial relation (NS)}, and \textbf{attribute binding (AB)}~\cite{dinh2022tise,Grimal_2024_tiam,gordon2023mismatch}. Usually, the prompt is composed of multiple distinct pieces of information that describe different parts of a scenery. These pieces of information accumulate into a rich description of the scene. Specifically, a complex prompt can be decomposed into a set of disjoint assertions that describe different parts of the content, \eg{} single or multiple objects, relations between objects, object attributes, lighting, style, and artistic reference. Thus, the composition of assertions must be known or extracted from the prompt. This concept is akin to Winoground's~\cite{Thrush_2022_CVPR} notion of visio-linguistic compositional reasoning. It refers to the task of understanding and reasoning about the relationships between visual and textual components in a way that requires combining them to form a coherent understanding or to make inferences. This involves tasks that require not just recognizing objects or elements in images and understanding text but also understanding how the textual and visual elements interact and convey a particular meaning.

In the category of image-text metrics, we have further identified the subcategories \textit{embedding-based metrics}, which quantify image generation quality based on text-image alignment \eg{} PickScore~\cite{Kirstain2023PickaPicAO}, ImageReward~\cite{xu2023imagereward}, Human Preference Score~\cite{Wu_2023_ICCV,Wu2023HumanPS}, and \textit{content-based metrics}, which examine the content of both the generated image and the text prompt. We will discuss both of these subcategories in more detail below.

\subsubsection{Embedding-based Metrics}\label{ssec:embedding-based}
\textit{Embedding-based metrics}, quality evaluation is based on learned embedding representations for vision and language inputs. Therefore, a text prompt $t$ is tokenized by a tokenizer and is then transformed into an embedding vector $f_t = F^{\theta}_T(t)$ using a text encoder model $F^{\theta}_T$, \eg{} a transformer~\cite{radford2019language}. Similarly, image $x$ is transformed into an image embedding representation $f_i = F^{\theta}_I(x)$ using an image encoder model $F^{\theta}_I$, \eg{} a ViT~\cite{dosovitskiy2021an}. These embedding vectors $f_t$ and $f_i$ have a fixed size and carry compressed information of both representations. Since the foundation models used are trained through representation learning in order to output meaningful embeddings, the cosine distance $D_{cos}$ between text and image embeddings can be computed to measure alignment:

\begin{equation}
    D_{cos}(f_t, f_i) = 1 - S_{cos}(f_t, f_i)
    \label{eq:cosine_distance}
\end{equation}
whereby the cosine similarity $S_{cos}$ is given by:

\begin{equation}
    S_{cos}(f_t, f_i) = \frac{f_i \cdot f_t}{\|f_i\| \|f_t\|}
    \label{eq:cosine_similarity}
\end{equation}

When training a powerful T2I model, embedding vectors for text-image pairs are aligned via vision-and-language pretraining strategies, \eg{} CLIP~\cite{radford2021learning}, BLIP~\cite{li2022blip}, or BLIP-2~\cite{Li2023BLIP2BL}. The embedded vectors extracted from these models encode valuable information, resulting in superior performance for multiple zero-shot scenarios~\cite{li2022blip,Li2023BLIP2BL,yuksekgonul2022and}. For example, the widely used CLIPScore~\cite{hessel2021clipscore} metric is defined by:

\begin{equation}
    QM_{clip}(t, x) = \omega * max(S_{cos}(F^{\theta}_T(t), F^{\theta}_I(x)), 0)
    \label{eq:clip-score}
\end{equation}
which re-scales the cosine similarity metric by the factor $\omega$.

However, several works have shown that pre-trained representations can be further fine-tuned on human-annotated data. In this way, human judgments can also be incorporated into the embeddings, as demonstrated by PickScore~\cite{Kirstain2023PickaPicAO} and ImageReward~\cite{xu2023imagereward}. Embedding-based metrics are computed utilizing these embedding representations, e.g., measuring similarity via \Cref{eq:cosine_similarity}, or regressing a score learned by a feed-forward network $F_{e}^{\theta}$:
\begin{equation}
    QM_{embedding}(t, x) = F_{e}^{\theta}(F^{\theta}_T(t), F^{\theta}_I(x))
\end{equation}

\subsubsection{Content-based Metrics}\label{ssec:content-based}
\textit{Content-based metrics} analyze language and visual representations with respect to their semantic content, whereby the actual measurements of such metrics are computed for decomposed components separately. Content-based metrics are based on the way in which humans would compare content across the text and image domains, \eg{} reading words in a prompt and matching them to regions depicted in the image, and vice versa. Hence, content-based quality metrics are comprehensible for human observers due to their relatable behavior, and are thus opposed to embedding-based metrics, inherently explainable.

\noindent\textbf{Text-Image Content Matching.} To relate parts of the text prompt to image regions, the text prompt needs to be dissected into substrings, where each substring describes distinct details, \eg{} an object, the relation between two objects, scene settings, \etc{} This decomposition into distinct semantic elements $S_T = \{ s_1, s_2, \ldots, s_n \}$ is elementary for text-image content matching. An automated process $\Phi_T$, usually based on an LLM, performs decomposition at the word level. Some benchmark datasets~\cite{Gokhale2022BenchmarkingSR} synthesize prompts using prompt templates to generate object relations, \eg{} ``\{objectA\} \{spatial relation\} \{objectB\}''. Using such a prompt dissection method $\Phi_T$, the resulting set of elements $S_T$ is compared to the corresponding regions in the image. This can be done using a visual question answering model (VQA)~\cite{Li2023BLIP2BL,openai2023gpt}, where questions are generated based on the elements in $S_T$. The VQA model $F^{\theta}$ is then interrogated for the presence or absence of specific relations, objects, attributes \etc{}, and the distance measure is computed by facilitating the VQA:
\begin{equation}
    QM^{F^{\theta}}_{TI}(t, x) = F^{\theta}(\Phi_T(t), x)
    \label{eq:vqa_qm}
\end{equation}
Another way to assess content alignment is to match elements from the prompt to regions of the image. Therefore, meaningful regions from the image using an image-based detector $\Phi_I$ are extracted resulting in candidate regions found in the image represented by a set of visual elements $S_I = \{ s_1, s_2, \ldots, s_m\}$. This is usually done through object detection~\cite{minderer2022simple} or semantic segmentation~\cite{reis2023real}. Then, a distance measure $D_{TI}$ computes matches elements from $S_T$ to elements inside $S_I$.

\begin{equation}
    QM^{\Phi_I}_{TI}(t, x) = D_{TI}(\Phi_T(t), \Phi_I(x))
\end{equation}

\noindent\textbf{Image-Text Content Matching.} In contrast to text-image content matching, image-text content matching starts the dissection on the image side. Thus, $\Phi_I$ derives a set of visual elements $S_I$ corresponding to regions in the image. Consequently, a distance measure $D_{IT}$ matches the image regions to the corresponding positions of the prompt. Image-text content matching can thus be formalized as follows:

\begin{equation}
    QM_{IT}(t, x) = D_{IT}(\Phi_T(t), \Phi_I(x))
\end{equation}

It is important to observe that the relation between text-image and image-text content matching is not bijective, since there might be parts in the image that are not mentioned in the text and vice versa, for example $S_T \cup S_I \neq S_T \cap S_I$. 

Instead of extracting visual elements directly from the image, an image caption model $\eta$ can also be used to generate captions~\cite{wu2022grit,openai2023gpt}, which then in turn describe the presented scene. Image captioning is an ongoing topic within the vision-language model community, and for the evaluation of such models, image caption metrics are utilized~\cite{Vedantam_2015_CVPR,10.1007/978-3-319-46454-1_24,Cui_2018_CVPR,jiang2019tiger,madhyastha-etal-2019-vifidel,lee-etal-2020-vilbertscore} and image-text matching can be based on an image caption metric $D_c$ as follows:

\begin{equation}
    QM^{\eta}_{IT}(t, x) = D_c(t, \eta(x))
\end{equation}

\section{Metrics}\label{sec:metrics}
In this section, we provide an overview of key metrics used to evaluate image quality in T2I generation systems. The assessment of image quality for T2I generation has evolved significantly to address the multimodal nature of the task, as opposed to traditional image quality metrics, which do not consider the text prompt $t$. Therefore, this section first reviews approaches considering image $x$ and text prompt $t$ simultaneously, before reviewing image-only metrics. We begin with embedding-based metrics (\Cref{ssec:metrics-embedding-based}), which leverage shared embedding spaces to assess the semantic alignment between text and images. Next, we explore content-based metrics (\Cref{ssec:metrics-content-based}), including methods for evaluating text-image content matching (\Cref{sssec:metrics-tti-metrics}) and image-text content matching (\Cref{sssec:metrics-itt-metrics}), focusing on the interplay between text descriptions and generated content. Finally, we discuss image-only metrics (\Cref{ssec:metrics-image-based}), which evaluate visual quality independently of textual information. This includes metrics that compare the distribution of generated images (\Cref{ssec:metrics-distribution-metrics}) and those that assess the quality of individual images (\Cref{ssec:metrics-single-image-metrics}). Together, these approaches provide a comprehensive framework for evaluating the multimodal and visual aspects of T2I generation.
To identify the relevant T2I quality metrics for this survey, we have performed a systematic literature study. We started our study with a selection of seminal papers from prominent conferences and journals, including CVPR, ICCV, ECCV, and NeurIPS. These publications were selected to ensure coverage of influential methodologies within the domain, which led us to the following seed papers: CIDERr\cite{Vedantam_2015_CVPR}, CLIPScore\cite{hessel2021clipscore}, PickScore\cite{Kirstain2023PickaPicAO}, Image Reward\cite{xu2023imagereward}, Human Preference Score\cite{Wu2023HumanPS}, Inception Score\cite{szegedy2016inception}, and Fréchet Inception Distance\cite{heusel2017gans}. To ensure our process was robust, we also employed a systematic exploration of recent proceedings from these conferences over the last years, where we manually examined their titles and abstracts for relevance to T2I quality evaluation metrics. This additional step helped capture impactful works that might not yet have reached high citation counts but are recognized as important by the research community. Leveraging ChatGPT-based research tools to efficiently navigate conference papers was instrumental in this process. The selection of seed papers was a critical step, as it formed the basis for the subsequent literature review process. Based on the identified seed papers, we expanded our corpus by exploring the citation tree of these using the Google Scholar database, enabling us to identify and incorporate additional relevant studies. During this step, we applied a systematic keyword-based search strategy to identify relevant works, using terms such as “T2I metrics,” “text-to-image quality evaluation,” “generative models evaluation,” and “image captioning metrics.” Recognizing the rapid pace of advancement in this field, during this step, we also included non-peer-reviewed preprints from arXiv in our analysis. We assume that excluding such works would risk omitting significant contributions that may be formally published at a later time. Within this section, we will present and discuss the papers which we have included in this survey.

\Cref{tab:text-image-alignment-metrics} presents an overview of the reviewed quality metrics, where we compare the metrics on four compositional aspects, \cf{}~\Cref{ssec:t2i-metrics} and award points accordingly. Zero points are awarded when the metric is neither text-conditioned nor designed to capture the corresponding aspect. One point is awarded when the metric is text-conditioned but is not explicitly trained to reflect compositionality. Two points are awarded for fine-tuning or other optimizations for certain aspects. The metrics to which we have assigned three points are specifically designed to reason about the aspect in question, such as object detection, segmentation, or dedicated visual question answering.

\begin{table*}[]
    \resizebox{1.00\linewidth}{!}{ 
    \begin{tabular}{cclccccccccc}
    \toprule
    &&                                                                &                     &         &      & \multicolumn{4}{c}{\cellcolor{graycolor}Compositional Ability}                       &      &    \\
    \multicolumn{2}{wc{5mm}}{\multirow{2}{*}{{Taxonomy}}}&\multirow{2}{*}{Metric}                                         &\multirow{2}{*}{Year}&Cites/ &Fine  & \textbf{Object}    & \textbf{Spatial}      &  \textbf{Non-Spatial} & \textbf{Attribute} & Human & \multirow{2}{*}{Rationale}    \\
    &&                                                                &                     &Year     &Tuned & \textbf{Accuracy}  & \textbf{Relations}    &  \textbf{Relations}   & \textbf{Binding}   & Evaluated &     \\
    \midrule                                                                                                                                                                                                   
    \multicolumn{2}{c}{\cellcolor{embcolor}} & CLIPScore~\cite{hessel2021clipscore}         &2021& 311  & \nej      & \oaO  & \srO & \nsO & \abO                 & \yay      & \nej     \\[-0.1mm]
    \multicolumn{2}{c}{\cellcolor{embcolor}} & BLIP-ITC~\cite{li2022blip}                   &2022& 1374 & \nej      & \oaO  & \srO & \nsO & \abO                 & \nej      & \nej     \\[-0.1mm]
    \multicolumn{2}{c}{\cellcolor{embcolor}} & BLIP-ITM~\cite{li2022blip}                   &2022& 1374 & \nej      & \oaO  & \srO & \nsO & \abO                 & \nej      & \nej     \\[-0.1mm]
    \multicolumn{2}{c}{\cellcolor{embcolor}} & BLIP2-ITC~\cite{Li2023BLIP2BL}               &2023& 2349 & \nej      & \oaO  & \srO & \nsO & \abO                 & \nej      & \nej     \\[-0.1mm]
    \multicolumn{2}{c}{\cellcolor{embcolor}} & BLIP2-ITM~\cite{Li2023BLIP2BL}               &2023& 2349 & \nej      & \oaO  & \srO & \nsO & \abO                 & \nej      & \nej     \\[-0.1mm]
    \multicolumn{2}{c}{\cellcolor{embcolor}} & MID~\cite{kim2022mutual}                     &2022& 10   & \nej      & \oaO  & \srO & \nsO & \abO                 & \yay      & \nej     \\[-0.1mm]
    \multicolumn{2}{c}{\cellcolor{embcolor}} & CLIP-R-Precision~\cite{park2021benchmark}    &2021& 32   & \yay      & \oaT  & \srT & \nsT & \abT                 & \yay      & \nej     \\[-0.1mm]
    \multicolumn{2}{c}{\cellcolor{embcolor}} & NegCLIP~\cite{yuksekgonul2022and}            &2022& 107  & \yay      & \oaT  & \srT & \nsT & \abT                 & \nej      & \nej     \\[-0.1mm]
    \multicolumn{2}{c}{\cellcolor{embcolor}} & MosaiCLIP~\cite{singh2023coarse}             &2023& 3    & \yay      & \oaT  & \srT & \nsT & \abT                 & \nej      & \nej     \\[-0.1mm]
    \multicolumn{2}{c}{\cellcolor{embcolor}} & CLoVe~\cite{castro2024clove}                 &2024& 3    & \yay      & \oaT  & \srT & \nsT & \abT                 & \nej      & \nej     \\[-0.1mm]
    \multicolumn{2}{c}{\cellcolor{embcolor}} & PickScore~\cite{Kirstain2023PickaPicAO}      &2023& 128  & \yay      & \oaT  & \srT & \nsT & \abT                 & \yay      & \nej     \\[-0.1mm]
    \multicolumn{2}{c}{\cellcolor{embcolor}} & ImageReward~\cite{xu2023imagereward}         &2023& 186  & \yay      & \oaT  & \srT & \nsT & \abT                 & \yay      & \nej     \\[-0.1mm]
    \multicolumn{2}{c}{\cellcolor{embcolor}} & HPSv1~\cite{Wu_2023_ICCV}                    &2023& 44   & \yay      & \oaT  & \srT & \nsT & \abT                 & \yay      & \nej     \\[-0.1mm]
    \multicolumn{2}{c}{\cellcolor{embcolor}} & HPSv2~\cite{Wu2023HumanPS}                   &2023& 80   & \yay      & \oaT  & \srT & \nsT & \abT                 & \yay      & \nej     \\[-0.1mm]
    \multicolumn{2}{c}{\cellcolor{embcolor}} & DreamSim~\cite{fu2023dreamsim}               &2023& 61   & \yay      & \oaT  & \srT & \nsT & \abT                 & \yay      & \nej     \\[-0.1mm]
    \multicolumn{2}{c}{\cellcolor{embcolor}} & COBRA~\cite{ma2024cobra}                     &2024& 2    & \yay      & \oaT  & \srT & \nsT & \abT                 & \yay      & \nej     \\[-0.1mm] 
    \multicolumn{2}{c}{\cellcolor{embcolor}} & R-Precision~\cite{8578241}                   &2018& 317  & \yay      & \oaT  & \srT & \nsT & \abT                 & \nej      & \nej     \\[-0.1mm]
    \multicolumn{2}{c}{\multirow{-17}{*}{\cellcolor{embcolor}\rotatebox[origin=c]{-90}{Embedding-based}}}
                                                                                            & RAHF~\cite{liang2023rich}  &2023& 27   & \yay     & \oaH  & \srT & \nsT & \abT   & \yay      & \img     \\[-0.1mm]
    \cellcolor{concolor}&\cellcolor{tticolor} & B-VQA~\cite{huang2023t2icompbench}             &2023&84    & \nej      & \oaZ & \srZ & \nsZ & \abH               & \yay      & \nej           \\[-0.1mm]
    \cellcolor{concolor}&\cellcolor{tticolor} & VISOR$_{cond}$~\cite{Gokhale2022BenchmarkingSR}&2022&22    & \nej      & \oaZ & \srT & \nsZ & \abZ               & \yay      & \img           \\[-0.1mm]
    \cellcolor{concolor}&\cellcolor{tticolor} & PA~\cite{dinh2022tise}                         &2022&6     & \nej      & \oaO & \srO & \nsZ & \abZ               & \yay      & \nej           \\[-0.1mm]
    \cellcolor{concolor}&\cellcolor{tticolor} & CA~\cite{dinh2022tise}                         &2022&6     & \nej      & \oaH & \srZ & \nsZ & \abZ               & \yay      & \nej           \\[-0.1mm]
    \cellcolor{concolor}&\cellcolor{tticolor} & SOA~\cite{hinz2020semantic}                    &2020&40    & \nej      & \oaH & \srZ & \nsZ & \abZ               & \yay      & \img           \\[-0.1mm]
    \cellcolor{concolor}&\cellcolor{tticolor} & VISOR~\cite{Gokhale2022BenchmarkingSR}         &2022&22    & \nej      & \oaH & \srT & \nsZ & \abZ               & \yay      & \img           \\[-0.1mm]
    \cellcolor{concolor}&\cellcolor{tticolor} & VISOR$_N$~\cite{Gokhale2022BenchmarkingSR}     &2022&22    & \nej      & \oaH & \srT & \nsZ & \abZ               & \yay      & \img           \\[-0.1mm]
    \cellcolor{concolor}&\cellcolor{tticolor} & TIAM~\cite{Grimal_2024_tiam}                   &2024&5     & \nej      & \oaH & \srZ & \nsZ & \abH               & \yay      & \img           \\[-0.1mm]
    \cellcolor{concolor}&\cellcolor{tticolor} & \textit{3-in-1}~\cite{huang2023t2icompbench}   &2023&84    & \nej      & \oaH & \srT & \nsZ & \abH               & \yay      & \img           \\[-0.1mm]
    \cellcolor{concolor}&\cellcolor{tticolor} & ViCE~\cite{betti2023let}                       &2023&4     & \nej      & \oaH & \srT & \nsH & \abH               & \yay      & \txt           \\[-0.1mm]
    \cellcolor{concolor}&\cellcolor{tticolor} & TIFA~\cite{Hu_2023_ICCV}                       &2023&83    & \nej      & \oaH & \srT & \nsH & \abH               & \yay      & \txt           \\[-0.1mm]
    \cellcolor{concolor}&\cellcolor{tticolor} & VNLI~\cite{yarom2023seetrue}                   &2023&31    & \yay      & \oaH & \srH & \nsH & \abH               & \yay      & \nej           \\[-0.1mm]
    \cellcolor{concolor}&\cellcolor{tticolor} & MQ~\cite{gordon2023mismatch}                   &2023&3     & \nej      & \oaH & \srH & \nsH & \abH               & \yay      & \txt + \img    \\[-0.1mm]
    \cellcolor{concolor}&\cellcolor{tticolor} & VQ$^2$~\cite{yarom2023seetrue}                 &2023&31    & \nej      & \oaH & \srH & \nsH & \abH               & \yay      & \txt           \\[-0.1mm]
    \cellcolor{concolor}&\cellcolor{tticolor} & VQAScore~\cite{lin2024evaluating}              &2024&57    & \yay      & \oaH & \srH & \nsH & \abH               & \yay      & \nej           \\[-0.1mm]
    \cellcolor{concolor}&\cellcolor{tticolor} & MINT-IQA~\cite{wang2024understanding}          &2024&5     & \yay      & \oaH & \srH & \nsH & \abH               & \yay      & \txt           \\[-0.1mm]
    \cellcolor{concolor}&\multirow{-16}{*}{\cellcolor{tticolor}\rotatebox[origin=c]{-90}{Text-Image}}
                                              & DA-Score~\cite{singh2023divide}                &2023&9     & \nej      & \oaH & \srH & \nsH & \abH               & \yay      & \txt           \\[-0.1mm]
    \cellcolor{concolor}&\cellcolor{ittcolor} & LEIC~\cite{Cui_2018_CVPR}                      &2018&27    & \nej      & \oaZ  & \srZ & \nsZ & \abZ              & \yay      & \nej      \\[-0.1mm]
    \cellcolor{concolor}&\cellcolor{ittcolor} & CIDEr~\cite{Vedantam_2015_CVPR}                &2015&544   & \nej      & \oaZ  & \srZ & \nsZ & \abZ              & \yay      & \nej      \\[-0.1mm]
    \cellcolor{concolor}&\cellcolor{ittcolor} & TIGEr~\cite{jiang2019tiger}                    &2019&13    & \nej      & \oaO  & \srO & \nsZ & \abO              & \yay      & \nej      \\[-0.1mm]
    \cellcolor{concolor}&\cellcolor{ittcolor} & SPICE~\cite{10.1007/978-3-319-46454-1_24}       &2016&257   & \nej      & \oaO  & \srO & \nsZ & \abO              & \yay      & \nej      \\[-0.1mm]
    \cellcolor{concolor}&\cellcolor{ittcolor} & T2T~\cite{chefer2023attend}                    &2023&218   & \nej      & \oaO  & \srO & \nsO & \abO              & \yay      & \nej      \\[-0.1mm]
    \cellcolor{concolor}&\cellcolor{ittcolor} & ViLBERTScore~\cite{lee-etal-2020-vilbertscore} &2020&11    & \nej      & \oaH  & \srT & \nsZ & \abZ              & \yay      & \nej      \\[-0.1mm]
    \cellcolor{concolor}&\cellcolor{ittcolor} & VIFIDEL~\cite{madhyastha-etal-2019-vifidel}    &2019&8     & \nej      & \oaH  & \srT & \nsZ & \abZ              & \yay      & \nej      \\[-0.1mm]
    \cellcolor{concolor}&\cellcolor{ittcolor} & UniDet~\cite{huang2023t2icompbench}            &2023&84    & \nej      & \oaH  & \srT & \nsZ & \abZ              & \yay      & \img      \\[-0.1mm]
    \cellcolor{concolor}&\cellcolor{ittcolor} & LLMScore~\cite{lu2023llmscore}                 &2023&33    & \nej      & \oaH  & \srT & \nsH & \abH              & \yay      & \txt      \\[-0.1mm]
    \multirow{-25}{*}{\cellcolor{concolor}\rotatebox[origin=c]{-90}{Content-based}}&\multirow{-10}{*}{\cellcolor{ittcolor}\rotatebox[origin=c]{-90}{Image-Text}}
                                              & VIEScore~\cite{ku2023viescore}                 &2023&16    & \nej      & \oaH  & \srT & \nsH & \abH              & \yay      & \txt      \\[-0.1mm]
    \multicolumn{2}{c}{\cellcolor{discolor}} & IS~\cite{NIPS2016_8a3363ab}                    &2016&1398 & \nej      &  \non & \non & \non & \non & \nej      & \nej     \\[-0.1mm]
    \multicolumn{2}{c}{\cellcolor{discolor}} & FID~\cite{heusel2017gans}                      &2017&1885 & \nej      &  \non & \non & \non & \non & \nej      & \nej     \\[-0.1mm]
    \multicolumn{2}{c}{\cellcolor{discolor}} & MiFID~\cite{bai2021mifid}                      &2021&11   & \nej      &  \non & \non & \non & \non & \nej      & \nej     \\[-0.1mm]
    \multicolumn{2}{c}{\cellcolor{discolor}} & KID~\cite{binkowski2018kid}                    &2018&241  & \nej      &  \non & \non & \non & \non & \nej      & \nej     \\[-0.1mm]
    \multicolumn{2}{c}{\cellcolor{discolor}} & C2ST~\cite{lopez2016revisiting}                &2016&55   & \nej      &  \non & \non & \non & \non & \nej      & \nej     \\[-0.1mm]
    \multicolumn{2}{c}{\cellcolor{discolor}} & PRD~\cite{sajjadi2018assessing}                &2018&93   & \nej      &  \non & \non & \non & \non & \nej      & \nej     \\[-0.1mm]
    \multicolumn{2}{c}{\cellcolor{discolor}} & CAS~\cite{ravuri2019classification}            &2019&44   & \nej      &  \non & \non & \non & \non & \nej      & \nej     \\[-0.1mm]
    \multicolumn{2}{c}{\cellcolor{discolor}} & DINO Metric~\cite{Ruiz_2023_CVPR}              &2023&1236 & \nej      &  \non & \non & \non & \non & \nej      & \nej     \\[-0.1mm]
    \multicolumn{2}{c}{\multirow{-9}{*}{\cellcolor{discolor}\rotatebox[origin=c]{-90}{Distribution}}}
                                             & I-PRD~\cite{kynkaanniemi2019improved}          &2019&139  & \nej      &  \non & \non & \non & \non & \nej      & \nej     \\[-0.1mm]
    \multicolumn{2}{c}{\cellcolor{sincolor}} & GMM-GIQA~\cite{gu2020giqa}                     &2020&19    & \nej      &  \non & \non & \non & \non & \nej      & \nej    \\[-0.1mm]
    \multicolumn{2}{c}{\cellcolor{sincolor}} & CLIP-IQA~\cite{wang2023clipiqa}                &2023&198   & \nej      &  \non & \non & \non & \non & \nej      & \nej     \\[-0.1mm]
    \multicolumn{2}{c}{\cellcolor{sincolor}} & Aesthetic Predictor~\cite{schuhmann2022laion}  &2022&999   & \yay      &  \non & \non & \non & \non & \yay      & \nej     \\[-0.1mm]
    \multicolumn{2}{c}{\cellcolor{sincolor}} & PAL4VST~\cite{zhang2023perceptual}             &2023&9     & \yay      &  \non & \non & \non & \non & \yay      & \img    \\[-0.1mm]
    \multicolumn{2}{c}{\cellcolor{sincolor}} & PAL4InPaint~\cite{zhang2022perceptual}         &2022&7     & \yay      &  \non & \non & \non & \non & \yay      & \img     \\[-0.1mm]
    \multicolumn{2}{c}{\cellcolor{sincolor}} & KPR~\cite{kynkaanniemi2019improved}            &2019&139   & \nej      &  \non & \non & \non & \non & \nej      & \nej     \\[-0.1mm]
    \multicolumn{2}{c}{\multirow{-7}{*}{\cellcolor{sincolor}\rotatebox[origin=c]{-90}{Single Image}}}
                                             & PPL~\cite{karras2019style}                     &2019&2188  & \nej      & \non & \non & \non & \non & \nej      & \nej     \\[-0.1mm]
    \midrule
    \multicolumn{9}{l}{{\txt: text-based rationale, \img: image-based rationale}} \\
    \bottomrule
    \end{tabular}} 
    \vspace{1pt} 
    \caption{Comparative overview of T2I evaluation metrics classified according to our proposed taxonomy, indicated by color: \textcolor{concolor}{blue} for text-conditioned metrics and \textcolor{discolor}{red} for image-only metrics. This table also categorizes current state-of-the-art methods based on their ability to assess compositional alignment, their validation through human evaluation studies, and their provision of additional rationale beyond a mere quality score.}
    \label{tab:text-image-alignment-metrics}
\end{table*}

\IEEEpubidadjcol

\subsection{Embedding-based Metrics}\label{ssec:metrics-embedding-based}
Text-conditioned image quality assessment represents a novel and evolving paradigm in the field of T2I generation. In these approaches, the perceived quality of an image is assessed not only based on its visual characteristics, but also in the context of accompanying textual information. However, existing image-only measures (\Cref{ssec:metrics-image-based}) are unable to integrate textual cues that describe the content associated with an image. T2I alignment acknowledges the intrinsic relationship between language and visual perception, allowing for a more nuanced evaluation that aligns with human judgment. In applications where text and image synergy is crucial, such as T2I generation, image captioning, content-based image retrieval, and human-computer interaction-based image generation, quantitatively measuring the alignment between text and image is mandatory. The incorporation of textual information introduces a dynamic dimension to image quality assessment, reflecting the evolving needs of multimodal systems and fostering advancements in the understanding and evaluation of visual content. In the following, we provide an overview of recent developments in quality assessment for T2I alignment. 

One of the first reference-free approaches, that were used for measuring the distance between a textual and an image representation is CLIPScore~\cite{hessel2021clipscore}, which is based on Contrastive Language-Image Pre-training (CLIP)~\cite{radford2021learning}. The CLIP distance is computed through the cosine similarity between the text embedding vector and the image embedding vector. By pre-training on vast and diverse datasets, CLIP exhibits a remarkable capacity to generate meaningful and contextually rich embeddings for images and corresponding textual descriptions. CLIPScore was introduced as a reference-free evaluation metric for image caption generation tasks together with its reference-based version RefCLIPScore. 

The Multimodal mixture of Encoder-Decoder (MED) was proposed by Li et al. \cite{li2022blip} and serves as a framework for multi-task pre-training and flexible transfer learning using image-text pairs from the web, integrated into BLIP, which supports various downstream tasks like T2I retrieval on datasets such as COCO Captions \cite{chen2015microsoft} and Flickr30K \cite{young2014flickr30k,plummer2015flickr30kentities}. BLIP employs image-text contrastive learning (ITC) to generate embedding vectors for computing cosine similarity, while its image-text matching (ITM) variant focuses on binary classification of image-text pairs. In 2023, BLIP2 was introduced as an efficient vision-language pre-training method that leverages pre-trained image encoders and large language models (LLMs) with fewer trainable parameters, achieving new benchmarks and advanced zero-shot capabilities for text generation from images. BLIP2 features a two-stage pre-training process with a query transformer (Q-Former) that enhances vision-language representation learning and generative learning, similarly utilizing learned embedding vectors to compute alignment scores (BLIP2-ITC) and offering an image-text matching version (BLIP2-ITM) as well.

Singh\etal\cite{singh2023coarse} additionally employs scene graphs and proposes a graph decomposition and augmentation framework to learn text-image representation. They derive a pseudo image scene graph from the text caption by dividing the text-based graph into multiple subgraphs and matching them with the image. They further extend the common vision-language component of the loss by an image-to-multi-text loss to train their model MosaiCLIP.


LXMERT, Tan\etal\cite{tan2019lxmert} introduced, is a transformer model employing three encoders for object relationships, language, and cross-modality, using text and image inputs and a separate object detection module for image encoding. This dependency is disadvantageous, as the object detector may miss some objects, possibly omitting vital information. On the other hand, the Unified Transformer (UniT) by Hu\etal\cite{Hu_2021_ICCV} simultaneously learns visual perception and language tasks without requiring task-specific tuning, handling text-image pairs with a task-specific index. UNITER by Chen\etal\cite{chen2020uniter} uses four pre-training tasks and also relies on an object detector, sharing LXMERT's drawback. Gan\etal\cite{gan2020large} enhance UNITER with large-scale task-agnostic adversarial pre-training and task-specific tuning, creating a more consistent embedding space. Zhang\etal\cite{Zhang_2021_CVPR} improve a vision-language framework by amplifying training data and refining the object detection model, aiding downstream representations, but unlike UniT, their method needs task-specific tuning. Kim\etal\cite{kim2021vilt} propose ViLT, a straightforward and effective vision-language pre-training approach. Unlike previous methods, their model doesn't use region features from an object detection model, instead applying a vision transformer encoder to generate patch-level representations, decreasing reliance on the detector for feature extraction. This also minimizes model size and boosts speed. The method uses a shared transformer encoder with modality tokens to distinguish between text and image inputs, along with specific token and patch positional embeddings. Li\etal\cite{Li_2019_ICCV} introduced the Visual Semantic Reasoning Network (VSRN), employing bottom-up attention to consider relevant image regions, further processed by a graph convolution network for semantic relationship features. The output is paired with text encoding, optimizing the encoding and text generation to align modalities together.

Another relevant metric for evaluating text-image alignment is R-Precision. Xu\etal\cite{8578241} were the first to apply it for text-image alignment. For this, they aim to identify the top $r$ relevant text captions for a given image with caption candidates $R$, to compute R-Precision as $r/R$. This is achieved by first extracting global feature vectors from their pre-trained encoders for generated images and given text captions. The cosine similarity is computed between image and text vectors, and then used to rank the captions in descending similarity to identify the $r$ most similar candidates. Park\etal\cite{park2021benchmark} extend this approach by using CLIP as the encoder for images and text, and show this leads to a more human-aligned judgment and prohibits the bias that might come from using a custom model. 

Kim\etal\cite{kim2022mutual} proposed Mutual Information Divergence (MID), a unified metric for multimodal generation, calculated through the negative Gaussian cross-mutual information between real and generated samples. Broadly formulated, their metric quantitatively measures how well one modality is aligned with the other, where both modalities are represented as encodings generated by their respective CLIP encoders. MID is shown to have consistent behavior across a variety of datasets where cosine-similarity-based techniques have shown weaknesses, especially for narrow domains like images of human faces. 
Kirstain\etal\cite{Kirstain2023PickaPicAO} developed a scoring function called PickScore to estimate user satisfaction with the generated images by fine-tuning CLIP-H on a large dataset of generated images and human preferences. Their objective maximizes the likelihood of a preferred image over an unpreferred one, culminating in a benchmark named Pick-a-Pic, which includes 500k examples and 35k distinct prompts that reflect human imagination in image generation. Xu\etal\cite{xu2023imagereward} advanced this concept by creating a T2I human preference reward model trained on 137k annotated text-image pairs, utilizing the model's output for Reward Feedback Learning to enhance a diffusion model's image generation, resulting in images more aligned with human preferences. Similarly, Wu et al. developed their human preference score (HPS) in two iterations; the first \cite{Wu_2023_ICCV} involved a dataset of 98k images and 25k prompts, where they fine-tuned a CLIP-L model to maximize similarity between prompts and user-chosen images while minimizing similarity to rejected images. The second version \cite{Wu2023HumanPS} expanded the dataset to 798,090 annotations for 433,760 image-text pairs, improving the scoring mechanism and demonstrating sensitivity to algorithmic enhancements in T2I models. Although these approaches improve the alignment with human preferences, they also highlight the challenges of accurately capturing subjective user satisfaction in image generation.

Fu\etal\cite{fu2023dreamsim} propose DreamSim, an extensive benchmark for the evaluation of generated images w.r.t. human preference alignment. Their dataset is composed of 20k synthetic image triplets with a reference image as well as two other images, where the user decided which is more similar to the reference. Their dataset covers various aspects of similarity, such as pose, perspective, foreground color, the number of items, and object shape. Using this dataset, they learn their perceptual metric using an ensemble of networks to encode each of the triplet images, calculate the cosine similarity between each image to the reference, followed by a triplet loss. They show their learned network is able to make more human-aligned judgments compared to \eg{} CLIP. On the other hand, previous methods did not rely on an ensemble configuration, increasing the computational cost of DreamSim.

In DreamBooth~\cite{Ruiz_2023_CVPR}, a combination of three metrics is used to evaluate the generation of multiviews of an object. To assess image quality, they compared a generated image with the ground truth image of the same view using the cosine similarity of CLIP~\cite{radford2021learning} and DINO~\cite{caron2021emerging}. Hereby, the CLIP-based metric only requires the images to show the same subject to return high similarities, whereas the DINO-based metric was included to measure more fine-grained differences. Lastly, they use the cosine similarity of CLIP embeddings of the generated image and the corresponding text prompt to measure prompt fidelity.


\subsection{Content-based Metrics}\label{ssec:metrics-content-based}

Content-based metrics evaluate the generated image directly based on its content, rather than the image's projection into an embedding space (see~\Cref{ssec:embedding-based}). 
This also allows for a decomposition of the evaluation of single aspects of the image quality like object accuracy (OA), spatial relationships (S), non-spatial relationships (NS) or attribute bindings (AB). 
In the following, we will discuss multiple content-based metrics.

\subsubsection{Text-Image Content Matching}\label{sssec:metrics-tti-metrics}
SeeTRUE(VNLI)\cite{yarom2023seetrue} is a metric that involves fine-tuning multimodal models such as BLIP2\cite{Li2023BLIP2BL} and PaLI-17B\cite{chen2022pali}, trained on 110K text-image pairs with binary alignment labels. It determines if an image "entails" a description with a "yes" or "no" answer, and a higher "yes" response rate implies stronger alignment. A key limitation is its black-box approach, making model refinement challenging and evaluation trust problematic. In response, Mismatch Quest\cite{gordon2023mismatch} offers an end-to-end trainable method with both visual and textual feedback in T2I models to pinpoint and clarify alignment issues. It produces a broad training set with both aligned and misaligned image-text pairs, employing LLMs, visual grounding models, and POS Tagging to synthesize misalignments. This TV-Feedback training set allows feedback models to provide visual (bounding box) and textual misalignment explanations. Evaluated on the SeeTRUE Feedback dataset with 2,008 human annotations, it aligns well with human assessments, though it may struggle with multiple misalignments.

In difference to the fine-tuned metrics of Mismatch Quest and SeeTRUE(VNLI), other metrics are utilizing VQA models to generate an evaluation score for generated images. They especially utilize these VQA models for evaluation of disjoint parts of the image prompt, making them belong to the category of compositional metrics.

The Decompositional-Alignment Score (DA-Score) by~\cite{singh2023divide} evaluates T2I alignment by breaking down image prompts to address the limitation that models like CLIP might overlook misalignments, especially with complex prompts. DA-Score divides prompts into separate assertions, assessed individually by a VQA model (BLIP), providing insights into the generative model's strengths and weaknesses and facilitating optimization in the diffusion process by modulating low-scoring assertions' cross-attention. The authors show DA-Score aligns better with human evaluations than metrics like CLIP~\cite{hessel2021clipscore}, BLIP~\cite{li2022blip}, and BLIP2~\cite{Li2023BLIP2BL}, notably for intricate prompts. Nonetheless, both SeeTRUE(VQ$2$) and DA-Score need precise prompt crafting to evaluate image aspects, highlighting the importance of choosing a suitable evaluation dataset.

Hinz\etal\cite{hinz2020semantic} propose the Semantic Object Alignment (SOA) metric, aimed at addressing challenges in complex and multi-object scenes in generated images. They employ a pre-trained detection model to find prompted objects in pictures, sampling captions from the COCO validation set that name one of 80 primary object categories. Their user study reveals that SOA closely matches human rankings, unlike metrics such as the Inception Score. Similarly, Grimal\etal\cite{Grimal_2024_tiam} introduce the Text-Image Alignment Metric (TIAM) for evaluating the alignment between images and prompts using a pre-trained segmentation model. They craft prompts using a template enhanced with word labels and optional attributes, assessing color attributes at a 40\% detection threshold within segmentation masks. Utilizing YoloV8, trained on 80 COCO classes, they recommend prompts beginning with \textit{"a photo of"} to ensure realistic image synthesis. Their findings show a stronger correlation with human evaluations compared to earlier studies\cite{radford2021learning} and\cite{li2022blip}. However, both studies encounter limitations when applied to generative models that produce diverse styles, such as cartoons or sketches, and are limited by the training classes. A notable issue is the potential overlap between models used for generation and evaluation, as a recent study\cite{dinh2022tise} points out, highlighting that SOA shares the same pre-trained detector with CPGAN\cite{liang2020cpgan}, leading to possible overfitting and biased evaluation. A suggested remedy is to change the detection model used during evaluation.

The authors of VISOR~\cite{Gokhale2022BenchmarkingSR} found that many existing models struggle with the challenge of generating multiple objects, and even when successful, they often fail to capture spatial relationships described in the input text prompts. They propose three variants of the VISOR metric: VISOR, VISOR$_N$, and VISOR${_{cond}}$. These metrics first rely on detecting objects that have been mentioned in the text prompt using a pre-trained object detector and the centroids of detected bounding boxes for deriving depicted relationships. The VISOR metric returns 1 if all objects are present in the image with correct spatial relationships; otherwise, it returns 0. VISOR$_N$ adopts a distribution-based approach, assessing the model's ability to generate at least $n$ spatially correct images based on the VISOR score for a given text prompt mentioning spatial relationships. Finally, VISOR${_{cond}}$ evaluates the conditional probability of generating correct spatial relationships, given that all objects are generated accurately. This means that object accuracy does not influence the VISOR${_cond}$ metric. 

Dinh \etal\cite{dinh2022tise} propose two metrics for evaluating T2I generation: Positional Alignment (PA) and Counting Alignment (CA). PA evaluates how generated images align with positional details in text by defining positional words ($W$) like "above" and "below." 
Meanwhile, CA assesses a T2I model's accuracy with counting details in text, focusing on object numbers in images. These metrics shed light on how well images align with text, despite some challenges in capturing all positional and counting subtleties. The authors suggest T2I evaluation should consider various factors. They propose a framework that combines multiple evaluation aspects, termed a "bag of metrics," which is shown to offer more consistent rankings with real images and human assessments. 

Similar to this approach,~\cite{huang2023t2icompbench} introduced the 3-in-1 metric for evaluating attribute bindings, spatial relationships, and non-spatial relationships like "look at," "hold," and "play with" in T2I models. This metric combines three evaluation criteria to thoroughly analyze image content. For attribute bindings, the "Disentangled BLIP-VQA" method is used because typical VQA assessments often misinterpret object-attribute links. It divides complex prompts into single attribute-object questions to avoid confusion in VQA. The UniDet model examines spatial relations such as "next to," "near," "on the side of," and directions like "left," "right," "top," and "bottom." Non-spatial relations are assessed with CLIPScore~\cite{hessel2021clipscore}, rounding out the 3-in-1 metric.

In their paper Yuksekgonul\etal\cite{yuksekgonul2022and}, the authors aim to elucidate how Visual Language Models (VLMs) encode the compositional relationship between objects and attributes. To achieve this goal, they introduce the Attribution, Relation, and Order benchmark. This benchmark evaluates the VLM's comprehension of object properties and relations using the Visual Genome Attribution and Visual Genome Relation datasets, respectively. They evaluated order sensitivity using COCO~\cite{lin2014microsoft} and Flicker30k~\cite{young2014flickr30k}. The authors emphasize a critical issue regarding contrastive pretraining in VLMs, which tends to prioritize learning low-level features over higher-level compositional structures. To address this challenge, the authors propose composition-aware hard negatives, which they integrate into CLIP's contrastive objective~\cite{radford2021learning}. These hard negatives are generated by altering linguistic elements such as nouns and phrases in negative captions. During training, when assembling a batch of images and their corresponding captions, the authors include not only the original images but also strong alternatives. 
Through their evaluations, the authors assert that integrating the proposed alternatives improves the comprehension of VLMs' composition and order.


The Visual Instruction-guided Explainable Score (VIEScore) proposed by Ku \etal\cite{ku2023viescore} is composed of the perceptual quality (PQ) and the semantic consistency (SC) score. Both scores are based on the instruction of an LLM using hand-crafted prompt templates to reason about a given image. 

LLMScore, introduced by Lu\etal\cite{lu2023llmscore}, is the pioneering method utilizing LLMs for automatic T2I evaluation, applied in an image-to-text way. Initially, BLIP2 is employed for image captioning, creating a broad image description, followed by local reasoning focused on objects. Grit~\cite{wu2022grit} identifies object crops within the image and provides a textual description for each region. GPT-4~\cite{openai2023gpt} then integrates the global and local text descriptions, developing an object-centric visual description. The LLMScore's evaluation aim can be redirected, as shown by Lu\etal, who illustrate both scoring and error-checking goals. The visual description and evaluation directive are sent to GPT-4, returning the final LLMScore with a rationale. However, captions made by LLMs might introduce extra details invented by the LLM itself, not generated from the image captioning, possibly causing incomplete integration of the original prompt's requirements and input.

Visual Concept Evaluation (ViCE) is a metric intended to mimic human-like comprehension of visual concepts, allowing for direct concept generation upon prompt inspection. Like other VQA approaches, ViCE utilizes GPT-3.5-turbo~\cite{openai2023gpt} to form question-answer pairs from prompts. It starts with 15 initial questions to an LLM to interpret visual concepts. This method uniquely allows the model to seek extra details for refining its image understanding. After the initial responses, the LLM iteratively inquires about further information until the model achieves a satisfactory comprehension of the image, thus verifying the semantic relationships of objects. The final analysis of the visual image is executed by a BLIP2-based VQA model, which evaluates the image using the prior question-answer pairs.

In the work of Hu\etal\cite{Hu_2023_ICCV} they propose a metric called TIFA that uses VQA models to measure the faithfulness of a generated image. To do so, they generate multiple-choice question-answer pairs utilizing GPT-3~\cite{brown2020language} via in-context learning and apply verification of the generated questions using a multitask question-answering model called UnifiedQA~\cite{khashabi2020unifiedqa}. TIFA adopts an open-domain pre-trained vision language model (the authors recommend using mPLUG-large~\cite{li-etal-2022-mplug}) as a VQA model, rather than closed-class classification models fine-tuned on VQAv2~\cite{Goyal_2017_CVPR} enabling it to perform well on a diverse set of visual elements. However, limitations of TIFA are the dependency on 12 categories: object, activity, animal, food, counting, color, material, spatial, location, shape, attribute, and other, which are considered to generate question-answer pairs.

In the work of Lin\etal\cite{lin2024evaluating} VQAScore is proposed where the input to the model is an image $I$ and a question $Q$ in the following format: "Does this figure show \{text\}? Please answer yes or no." where \{text\} is the prompt used to generate the image $I$. They fine-tuned a VQA model to predict the answer likelihoods adopting a pre-trained bidirectional encoder-decoder language model, FlanT5~\cite{chung2022scaling} and combined it with a pre-trained CLIP vision encoder. In their evaluation on several alignment benchmarks~\cite{Thrush_2022_CVPR,Hu_2023_ICCV,Kirstain2023PickaPicAO,saharia2022photorealistic,ho2022imagen} they outperform models trained with extensive human feedback and divide-and-conquer methods. However, since VQAScore outputs the probability $P(Yes|I,Q)$, it does not provide any reasoning accompanying the predicted score.

The MINT-IQA (Multimodal INstruction Tuning Image Quality Assessment) model proposed by Wang\etal\cite{wang2024understanding} evaluates and explains human preferences for the generation of text-conditioned images in multiple dimensions such as quality, authenticity, and text-image correspondence. The model utilizes a vision-language instruction tuning approach, allowing for a deeper understanding and a more comprehensive evaluation of human visual preferences. Extensive experiments show that MINT-IQA achieves state-of-the-art performance on both AI-generated and traditional image quality assessment databases, underscoring its adaptability and the breadth of its applicative power.


\subsubsection{Image-Text Content Matching}\label{sssec:metrics-itt-metrics}
To assess T2I alignment, it is essential to consider its inversion. Prior to the development of CLIP, T2I was evaluated inversely, through image-to-text alignment. This approach is key in image caption or description generation tasks, where the task is to assess the textual output for a provided image. Datasets like Flickr8K~\cite{hodosh2013framing}, Flickr30K~\cite{young2014flickr30k}, MS-COCO~\cite{lin2014microsoft,chen2015microsoft}, and Pascal 50S~\cite{Vedantam_2015_CVPR} offer human assessments of captions for given images, and serve as benchmark datasets for image-to-text evaluation, leading to the development of various visio-linguistic metrics. Below, we outline the image captioning and machine translation metrics that influenced text-conditioned image generation evaluation before the advent of compositional quality metrics.

The SPICE metric (Semantic Propositional Image Caption Evaluation)\cite{10.1007/978-3-319-46454-1_24} evaluates the semantic details of text generated for image captions by converting both generated and reference sentences into scene graphs of objects, attributes, and relationships, comparing them using an F-score. Conversely, the LEIC metric (Learning to Evaluate Image Captioning) by Cui\etal\cite{Cui_2018_CVPR} leverages a CNN for image coding and an LSTM for text coding, utilizing a binary classifier to compare generated text quality with human judgment, potentially mirroring human assessment more closely than conventional metrics. TIGEr (Text-to-Image Grounding for Image Caption Evaluation)\cite{jiang2019tiger} improves evaluations by integrating text-image grounding to consider image content, showing better alignment with human judgment than word-based metrics such as \textsc{Bleu}, ROUGE, and METEOR. Furthermore, VIFIDEL\cite{madhyastha-etal-2019-vifidel} assesses visual fidelity by matching objects detected in images with their textual descriptions using word's mover distance (WMD), translating it into a similarity measure to check the match between object types and descriptive words, allowing for object priority based on word frequency in reference texts.

In the work of Lee\etal\cite{lee-etal-2020-vilbertscore} they propose a metric called ViLBERTScore, which is similar to BERTScore~\cite{Zhang2020BERTScore} that computes textual embeddings for a reference and a generated caption. Additionally, the computation of textual embeddings is conditioned on the target image using the model proposed by Lu\etal\cite{NEURIPS2019_c74d97b0}. Hereby, contextual embeddings are computed by applying an object detector to the target image and feeding pairs of image region features and text embeddings to the pre-trained ViLBERT model. Finally, the ViLBERTScore is defined by the cosine similarity between reference caption embeddings and candidate caption embeddings.

Common metrics for machine translation and image captioning model evaluation are utilized in text-image retrieval tasks. The CIDEr score~\cite{Vedantam_2015_CVPR} assesses how closely a generated sentence matches a set of human-written references for an image. This involves TF-IDF weighting to highlight distinctive n-grams and calculating cosine similarity over different n-gram lengths to produce a normalized score. \textsc{Bleu}~\cite{10.3115/1073083.1073135} calculates machine translation quality via $n$-gram precision between a candidate and human translations, with values from $0$ to $1$—a perfect score of $1$ is rare, yet it is favored for its alignment with human evaluations and efficiency. ROUGE~\cite{lin-2004-rouge}, initially for text summaries, evaluates recall with n-grams, incorporating longest sequences and skip-bigram co-occurrence, aligning well with human preferences. METEOR~\cite{banerjee-lavie-2005-meteor} enhances \textsc{Bleu} by integrating precision and recall through unigram matches, including stems and synonyms, offering better human judgment correlations. Finally, BertScore~\cite{Zhang2020BERTScore}, leveraging BERT~\cite{devlin2018bert}, measures textual quality via cosine similarity of word embeddings, effectively capturing semantics and context, and showing strong human judgment alignment.


\subsection{Image-Only Quality Metrics}\label{ssec:metrics-image-based}
\subsubsection{Distribution Metrics}\label{ssec:metrics-distribution-metrics}
A set of popular evaluation metrics assumes that the generative model is a black box and operates only on samples of the generated distribution $q$ and compares it with samples of the target distribution $p$. The most commonly used metrics then rely on comparing features produced by pre-trained neural networks.
Inception Score (IS)~\cite{NIPS2016_8a3363ab} uses an Inception network pre-trained on ImageNet to compare class predictions for a set of generated samples $x\sim q$. Here, the score rewards low entropy in class predictions $p(y|x)$, \ie{} generated images that can be clearly classified as one of the classes, as well as high entropy in marginal class distribution $p(y)$, \ie{} a large diversity among generated samples. Due to its short-comings~\cite{barratt2018note}, the IS has recently lost popularity. The MODE score~\cite{che2016mode} improves the IS by adding another term that rewards a similar distribution of class predictions for the generated and target images. IS-based metrics are not suitable for T2I, as the marginal class distribution $p(y)$ is typically not available.

The Fréchet Inception Distance~(FID)~\cite{heusel2017gans} compares the means and co-variances of the features, extracted by the Inception network from samples of the generated and target distributions, using the Fréchet distance (or Wasserstein-2). FID was shown to be a more consistent quality measure than IS and is still widely used. 
MiFID~\cite{bai2021mifid} extends FID by incorporating a term that penalizes the memorization of training samples, by computing the minimum cosine distance of Inception features to the training dataset. This penalty was introduced to avoid bogus submissions for an image generation competition.

All Inception-based metrics share the downside of relying on the weights of the Inception network. Those weights are the result of supervised ImageNet classification training, and many of the Inception metrics are not robust to different sets of weights obtained from similar trainings~\cite{barratt2018note}. Furthermore, with the increasing scale of modern T2I models and datasets~\cite{schuhmann2022laion} far beyond the ImageNet domain, the features trained to classify this comparatively narrow domain may be insufficient for quality assessment. 
Adoption of a more capable and general feature extractor, such as semi-supervised models, could improve the reliability of metrics like FID, especially for models exceeding the ImageNet domain.

Distributions $p$ and $q$ can be compared using kernel embedding methods, notably the maximum mean discrepancy (MMD) metric, which quantifies the distance between kernel embeddings of samples without the need for density estimation or bias correction, a significant advantage over information-theoretic approaches. However, the reliance on a fixed kernel in MMD can lead to issues when dealing with complex natural images. The Parzen window estimate~\cite{gretton2006kernel} is an example of an MMD approach, while the Kernel Inception Distance (KID)\cite{binkowski2018kid} improves upon this by calculating the squared MMD between Inception representations, thus addressing the bias present in the Fréchet Inception Distance (FID) related to sample size. Another method for comparing distributions is through two-sample tests, such as the C2ST introduced by Lopez-Paz and Oquab\cite{lopez2016revisiting}, which employs a binary classifier to differentiate between samples from the generated and target distributions, aiming for approximately 50\% accuracy with large sample sizes. This approach can be enhanced using a nearest-neighbor classifier, providing insights into the generated data; for instance, a predominance of generated images among the nearest neighbors may indicate mode collapse. C2ST is versatile, applicable to both nearest-neighbor and neural network-based classifiers, including those using pre-trained feature extractors such as ResNet-34~\cite{he2016deep}.

The previously introduced image-based metrics quantify image generation quality with a scalar score. Sajjadi\etal\cite{sajjadi2018assessing} introduce precision and recall for distributions (PRD), where precision is the proportion of generated images in the target distribution $p$, and recall is the proportion of real images in the generated distribution $q$. They analyze Inception embeddings of $p$ and $q$ with clustering of k-means and comparison of histograms. Clusters dominated by generated or target distribution samples affect precision and recall, respectively. They compute these metrics using multiple randomized clusterings. Kynkäänniemi\etal\cite{kynkaanniemi2019improved} enhance this method (I-PRD) by modeling the support manifold using hyperspheres around each embedding sample, allowing direct computation of precision and recall.

In the work of Ravuri\etal\cite{ravuri2019classification} the Classification Accuracy Score (CAS) is proposed. It is based on predictions for real images of a ResNet image classification model trained on synthetic data. The performance accuracy for the set of real images is referred to as CAS, and it is demonstrated that CAS can identify classes for which a GAN failed to correctly learn its data distribution.

\subsubsection{Single Image Quality Metrics}\label{ssec:metrics-single-image-metrics}
Gu\etal\cite{gu2020giqa} proposes GMM-GIQA, which models the embeddings of the target distribution $p$ using a Gaussian mixture model. A generated image can then be assigned a score based on the probability density of its embedding. The authors note, however, that the metric may fail for too complex distributions, as they cannot be sufficiently modeled using a Gaussian mixture model.

With CLIP at the center, Wang\etal\cite{wang2023clipiqa} proposes the CLIP Image Quality Assessment (CLIP-IQA) benchmark. In their work, they improve CLIP's ability to assess text-image alignment through antonym prompt pairing and removing the positional embedding from the image encoder. The resulting model is significantly better for evaluating quality and abstract perception. 

With the introduction of the LAION Aesthetics dataset~\cite{schuhmann2022laion}, the authors trained models\footnote{\scriptsize\url{https://github.com/christophschuhmann/improved-aesthetic-predictor}} to predict how aesthetic humans would rate a given generated image, resulting in an image quality metric that is aligned with human preferences.

Zhang\etal\cite{zhang2023perceptual} collects a dataset containing human-annotated segmentation of artifacts. They then train binary segmentation models to automatically detect such artifacts in images. They also propose a related metric to evaluate in-painting using generative models, the Perceptual Artifact Ratio (PAR)~\cite{zhang2022perceptual}, also known as PAL4InPainting, which measures the relative area occupied by artifacts. This metric is also generally applicable to full images, not just regions for in-painting.

Since the I-PRD method yields only a binary result for an individual sample, Kynkäänniemi\etal\cite{kynkaanniemi2019improved} proposes a variant, KPR, which estimates how close the feature vector of a single image is to the feature vectors of k-NN real images.

Karras\etal\cite{karras2019style} introduces the perceptual path length, a metric for latent variable models. The idea is to pairwise compare subsequent images in a latent space interpolation using a perceptual image quality metric. This metric measures whether any drastic changes appear for close latent codes and rewards smooth transitions within the interpolation, which is an indicator of good disentanglement. 




\section{Datasets}\label{sec:datasets}
This section provides an overview of datasets used to evaluate text-conditioned image generation, see \Cref{tab:datasets}. First, we compare datasets that originated from the image captioning research community, which were first used to evaluate text-image generation systems. With recent developments of vision-language models, researchers started to seek increased complexity of evaluation data resulting in the emergence of the term visiolinguistic compositionality. It describes the task and datasets to evaluate the ability of vision and language models to conduct reasoning of image and text that are subject to compositionality, meaning that they are ensembles of several contents. Second, we compare existing visual question answering (VQA) benchmark datasets, and finally, we list specifically designed datasets for the development of text-image quality metrics and their verification on human judgments.

The ranking system in \Cref{tab:datasets} evaluates prompts based on their source and complexity, assigning points from zero to three. Zero points are given for basic object, relation, and attribute labels, providing minimal information. One point is awarded to prompts derived from web scraping, offering a bit more context. Two points go to the prompts obtained through crowd sourcing, reflecting a higher level of detail and relevance. The highest score, three points, is reserved for prompts that accurately reflect the actual compositional intentions behind an image, showcasing the deepest understanding and context.

\begin{table*}[]
\resizebox{1.00\linewidth}{!}{
\begin{tabular}{lcccclcccc}
\toprule
\multirow{3}{*}{Dataset}                    & \multirow{3}{*}{Year}           & \multirow{3}{*}{Cites/Year}              & Number of         & Avg.   & \multirow{3}{*}{Source}                        & \multicolumn{4}{c}{\cellcolor{graycolor}Text Compositionality}                                 \\ 
                                            &                                  &                                           & Images /          & Number &                                               & Number           & Spatial          & Non-Spatial       & Attribute        \\ 
                                            &                                  &                                           & Text-Image Pairs  & Words  &                                               & Objects          & Relations        & Relations         & Binding          \\ 
\midrule
\multicolumn{10}{c}{Image Captioning} \\
\midrule
SBU Captioned Photo Dataset~\cite{NIPS2011_5dd9db5e} & 2011 & 118 & $1{,}000{,}000$ &       & Flickr.com                   & \oaO & \srO & \nsO & \abO \\
Pinerest40M~\cite{mao2016training}                  & 2016 & 6   & $40{,}000{,}000$& 10    & Pinerest.com                & \oaO & \srO & \nsO & \abO \\
Conceptual Captions~\cite{sharma2018conceptual}     & 2018 & 381 & $3{,}369{,}218$ & 10.3  & World Wide Web              & \oaO & \srO & \nsO & \abO \\
nocaps~\cite{Agrawal_2019_ICCV}                     & 2019 & 59  & $15{,}100$      & 10    & Open Images V4 (Flickr.com) & \oaO & \srO & \nsO & \abO \\
Conceptual 12M~\cite{Changpinyo_2021_CVPR}          & 2021 & 266 & $12{,}423{,}374$& 20.2  & World Wide Web              & \oaO & \srO & \nsO & \abO \\
UIUC Pascal Sentence Dataset~\cite{rashtchian2010collecting} & 2010 & 61   & $1{,}000$       & n.a.  & VOC2008                     & \oaT & \srT & \nsT & \abT \\
Flickr8K~\cite{hodosh2013framing}                   & 2013 & 134 & $8{,}092$       & n.a.  & Flickr.com                  & \oaT & \srT & \nsT & \abT \\
Flickr30K~\cite{young2014flickr30k}                 & 2014 & 265 & $31{,}783$      & n.a.  & Flickr.com                  & \oaT & \srT & \nsT & \abT \\
COCO Captions~\cite{chen2015microsoft}              & 2015 & 284 & $204{,}721$     & 11    & Flickr.com                  & \oaT & \srT & \nsT & \abT \\
PASCAL-50S~\cite{Vedantam_2015_CVPR}                & 2015 & 544 & $1{,}000$       & 8.8   & VOC2008                     & \oaT & \srT & \nsT & \abT \\
ABSTRACT-50S~\cite{Vedantam_2015_CVPR}              & 2015 & 544 & $500$           & 10.59 & ASD~\cite{Zitnick_2013_CVPR}& \oaT & \srT & \nsT & \abT \\
\midrule
\multicolumn{10}{c}{Visual Question Answering} \\
\midrule
VQA~\cite{Antol_2015_ICCV}                          & 2015 & 668 & $254{,}721$     & $<2$  & MS COCO \& Abstract Images  & \oaT & \srT & \nsT & \abT \\
VQAv2.0~\cite{Goyal_2017_CVPR}                      & 2017 & 425 & $204{,}721$     & n.a.  & MS COCO                     & \oaT & \srT & \nsT & \abT \\
VCR~\cite{Zellers_2019_CVPR}                        & 2019 & 167 & $110{,}000$     & 11.8  & LSMDC~\cite{rohrbach2016movie} \& YT & \oaT & \srT & \nsT & \abT \\
\midrule
\multicolumn{10}{c}{Compositionality Benchmarks} \\
\midrule
DrawBench~\cite{saharia2022photorealistic}          & 2022 & 1859 & $200$           & 11.69 & DALL-E, \cite{marcus2022preliminary}, Reddit & \oaH & \srH & \nsH & \abH \\
PaintSkills~\cite{Cho_2023_ICCV}                    & 2023 & 78   & $65{,}535$      & n.a.  & synthetic prompts           & \oaH & \srH & \nsH & \abH \\
ABC-6K~\cite{feng2023trainingfree}                  & 2022 & 96   & $6{,}400$       & n.a.  & MS COCO                     & \oaO & \srO & \nsO & \abH \\
CC-500~\cite{feng2023trainingfree}                  & 2022 & 96   & $500$           & n.a.  & synthetic prompts           & \oaH & \srZ & \nsZ & \abH \\
I2P~\cite{Schramowski_2023_CVPR}                    & 2023 & 117  & $4{,}703$       & 20.56 & user generated prompts      & \oaH & \srH & \nsH & \abH \\
Visual Genome~\cite{Krishna2016VisualGC}            & 2016  & 696  & $108{,}077$     & n.a.  & MS COCO                     & \oaH & \srH & \nsH & \abH \\
Winoground~\cite{Thrush_2022_CVPR}                  & 2022 & 129  & $800$           & 8.99  & Getty Images API            & \oaH & \srH & \nsH & \abH \\
RichHF-18K~\cite{liang2023rich}                     & 2024 & 54   & $18{,}000$      & n.a.  & Pick-a-Pic~\cite{Kirstain2023PickaPicAO} & \oaH & \srH & \nsH & \abH \\
T2I-CompBench~\cite{huang2023t2icompbench}          & 2023 & 83   & $6{,}000$       & 8.98  & generated prompts by GPT~\cite{openai2023gpt} & \oaH & \srH & \nsH & \abH \\
\bottomrule
\end{tabular}
}
\vspace{1pt} 
\caption{Comparison of text-image datasets based on the number of prompts, prompt length, compositional aspects of the prompts, and the context of the dataset provided in their textual data.}
\label{tab:datasets}
\end{table*}

\subsection{Image Caption Datasets}\label{ssec:image-caption-datasets}
The development of image generators requires tremendous amounts of image data~\cite{brock2018large,Karras_2020_CVPR} in order to learn data statistics and fit the output distribution of the generator to real image distributions. In the context of the evaluation of T2I generation, the necessity of text-image pairs arises. Fortunately, there already exist such datasets collected from researchers in the image captioning research domain, e.g., MS-COCO Captions~\cite{chen2015microsoft}, Flickr30K~\cite{young2014flickr30k,plummer2015flickr30kentities}, PASCAL-50S~\cite{Vedantam_2015_CVPR}, Abstrac-50S~\cite{Vedantam_2015_CVPR}, which curate one to fifty human-generated descriptions per image.

The UIUC Pascal Sentence Dataset~\cite{rashtchian2010collecting} and Flickr8K~\cite{hodosh2013framing} are among the first well-known image caption datasets, each providing multiple descriptions per image. Hodosh \etal\cite{hodosh2013framing} approach image description evaluation as a ranking task, including a collection of $8,092$ images from Flickr and $1,000$ from PASCAL VOC-2008~\cite{pascal-voc-2008}, each described by human annotators. Rashtchian \etal\cite{rashtchian2010collecting}'s crowdsourcing methodologies were used, collecting five descriptions per image through Amazon Turk. Participants generated single-sentence descriptions, focusing on central characters, settings, and object relations, using adjectives for attributes like color or emotion, in fewer than 100 characters. Another user group checked spelling and grammar to ensure high-quality descriptions. Later, Young \etal\cite{young2014flickr30k} expanded the Flickr8K dataset to $158,915$ captions covering $31,783$ images, and called it Flickr30K. Further, Plummer \etal\cite{plummer2015flickr30kentities} proposed extensions involving cross-caption coreference chains linking the same entities across image captions, with bounding boxes localizing these entities.

The SBU Captioned Photo Dataset~\cite{NIPS2011_5dd9db5e} collects one million images from Flickr.com ensuring some quality requirements; in particular, they filter the collected data for textual descriptions with a satisfactory length of visual description, at least two words belonging to (objects, attributes, actions, stuff, and scenes) and at least one preposition indicating visible spatial relation. While the dataset poses a tremendous amount of image-text pairs, the content of image captions may be visually descriptive but lacks human-supervised verification, resulting in many image captions being only comprehensible with personal knowledge of the caption's author, e.g., using the given name of a dog for describing a dog playing with a ball.

Based on Microsoft COCO~\cite{lin2014microsoft}, which is a large-scale dataset consisting of images acquired through Flickr showing multiple objects in their natural context, the frequently used COCO Captions~\cite{chen2015microsoft} dataset was created. It supplements MS-COCO by collecting $1,026,459$ captions for $164,062$ images, including five captions for each image in MS-COCO and a subset of $5,000$ images that were annotated with 40 reference sentences. Together with the actual dataset, the authors released an evaluation protocol; in particular, they deploy an evaluation server ensuring consistent evaluation computing numerous metrics like BLEU~\cite{10.3115/1073083.1073135}, ROGUE~\cite{lin-2004-rouge}, METEOR~\cite{banerjee-lavie-2005-meteor} and CIDEr~\cite{Vedantam_2015_CVPR}.

In the work of Vedantam\etal\cite{Vedantam_2015_CVPR} two datasets are collected, PASCAL-50S and ABSTRACT-50S based on the UIUC Pascal Sentence Dataset and the Abstract Scenes Dataset~\cite{Zitnick_2013_CVPR}, respectively. Annotations for these datasets were collected with the goal to investigate consensus between human annotators, in particular the similarity between a candidate image description and several reference descriptions. While PASCAL-50S features real images, ABSTRACT-50S consists of images in a clip-art style designed by humans in a different crowdsourcing study~\cite{Zitnick_2013_CVPR}. For both datasets, $50$ human-generated sentences are collected while annotators are instructed to provide descriptions that should help others recognize the image from a collection of similar images. Having a large set of fifty reference sentences per image facilitates research on text-image alignment; however, the amount and variety of images provided by both datasets seem too few in order to provide complexity for profound text-image evaluation.

Increasing the number of object classes is achieved by the image dataset called, nocaps. It consists of over $600$ object classes, and it is presented in the work of Agrawal\etal\cite{Agrawal_2019_ICCV}, which is based on OpenImages V4~\cite{kuznetsova2020open} a large-scale human-annotated object detection dataset. Nocaps was acquired by filtering Open Images and excluding images with non-zero or unknown image rotations, instances from a single object category, less than six unique object classes, and finally they apply a balancing scheme to have an even distribution of images depicting two to six unique object classes, avoiding frequently occurring object classes.

Mao et al.'s~\cite{mao2016training} introduction of the Pinterest40M dataset represents a significant advancement in multimodal word embeddings, featuring over 40 million images and 300 million sentences from Pinterest.com. Far exceeding the scale of existing datasets like MS COCO, Pinterest40M's unique blend of visual and textual data enables the development of richer word embeddings. Further, this dataset serves as a vital resource for exploring vision-language pre-training methods~\cite{kiros2018illustrative,jenkins2019unsupervised,Desai_2021_CVPR}.

The Conceptual Captions~\cite{sharma2018conceptual} dataset is derived from automated web crawling. This enables the collection of numerous image-text pairs, but stringent filtering is crucial to retain only high-quality content. Images are eliminated based on encoding, dimensions, aspect ratio, and inappropriate content. Given that Alt-text from HTML pages may lack detailed accuracy, it is refined using part-of-speech, sentiment, and inappropriate annotation analyses via the Google Cloud Natural Language APIs. For improved text quality, criteria such as noun and preposition frequency, token repetition, capitalization, English Wikipedia token likelihood, and known prefixes like "click to enlarge picture" or "stock photo" are applied. Image-text filtering with Google Cloud Vision APIs is conducted, matching text tokens with image content and replacing proper names with hypernyms through hypernymization. This dataset, containing over 3 million image-text pairs, is intended to support diverse downstream image captioning tasks but is predominantly used for vision-language pre-training~\cite{Changpinyo_2021_CVPR}.

To facilitate image-text pre-training, Conceptual 12M~\cite{Changpinyo_2021_CVPR} was acquired by relaxing filter criteria of the collection pipeline used for Conceptual Captions. This strategy trades precision of image descriptions for increased scale of the data corpus; in particular, they increase the recall of visual concept descriptions by lowering requirements of word repetitions, caption size ranges, image aspect ratios, and hypernymization. Just as Conceptual Captions and Pinterest40M, such web-sourced image descriptions enable vision-language pre-training but lack the quality and complexity for proper evaluation of T2I generation methods.



 \subsection{Visual Question Answering}\label{ssec:vqa}

The Visual Question Answering (VQA) dataset~\cite{Antol_2015_ICCV} is a groundbreaking tool, merging 123,287 MS COCO images and 50,000 abstract scenes with over 760,000 questions and 10 million answers collected via Amazon Mechanical Turk. This resource tests VQA models' abilities to interpret complex visual inputs, featuring a broad array of questions and answers reflecting real-world linguistic and visual diversity. For each image, five open-ended questions are presented, necessitating sophisticated visual recognition, commonsense understanding, and inferential thinking, with ten possible answers per question to encompass the range of human responses. VQA v2.0~\cite{Goyal_2017_CVPR}, an enhancement of VQA, addresses this by adding complementary images per question, forming question-image pairs with two distinct answers each. By doubling the dataset size, Goyal et al. tackle the issue of VLMs neglecting visual cues, crafting a model that can answer an image-question pair while providing a counterexample-based explanation.

The Visual Commonsense Reasoning (VCR) dataset~\cite{Zellers_2019_CVPR} is specifically designed to move beyond mere recognition tasks to test models on cognition-level visual understanding. It features 290,000 question-answer-rationale (QAR) triples across 110,000 unique movie scenes. Each QAR triple challenges models to not only identify objects within a scene but also to understand complex interactions and motivations. The dataset focuses on deep visual comprehension, requiring models to infer and rationalize about unseen aspects of the image, thus bridging the gap between visual perception and commonsense reasoning. VCR is frequently used as a downstream task for evaluating representation learning of visual-linguistic approaches~\cite{Su2020VL-BERT,lee-etal-2020-vilbertscore,chen2020uniter,pmlr-v139-cho21a}.




\subsection{Compositionality Benchmarks}\label{ssec:benchmark-datasets}

Thrush et al.\cite{Thrush_2022_CVPR} introduce Winoground, a new task and dataset for assessing vision and language models in visio-linguistic compositional reasoning. It requires matching two images with two captions, each with the same words in a different order, demanding precise modal understanding. Winoground, from the Getty Image API, involves human annotators creating creative captions and choosing corresponding images, tagging visual reasoning into object, relation, or both swaps. Winoground includes 1,600 pairs (400 examples), with 800 correct and 800 incorrect, featuring 800 unique images and captions. It prioritizes expert-quality annotations, serving as a probing dataset for linguistic and visual analysis.

The Visual Genome dataset~\cite{Krishna2016VisualGC} is a dataset for comprehensive scene understanding. It contains more than 108k images, each with an average of 35 objects delineated by a bounding box. However, bounding box annotated objects are not sufficient for comprehensive scene understanding. Object attributes and their relationships are also needed. To obtain these, about 50 overlapping image sub-regions per image have been captured by human annotators. From those, object attributes and relationships could be extracted, which in turn are used to create image scene graphs and additional question answer pairs. Object attributes and relationships are canonicalized on WordNet synsets~\cite{miller1995Wordnet}. 

T2I-CompBench~\cite{huang2023t2icompbench} is a compositional dataset targeting to provide complex prompt compositions in order to study attribute binding, object relations, and complex composition skills of image generation models. Therefore, they acquire a dataset consisting of 6,000 text-image pairs (1,000 for each sub-category: color, shape, texture, spatial relation, non-spatial relation, complex composition). Text prompts for color attribute binding are gathered from CC500~\cite{feng2023trainingfree} and COCO~\cite{chen2015microsoft}, while for the remaining sub-classes prompts are generated by GPT\cite{openai2023gpt} or handcrafted using prompt templates.

RickHF-18K dataset~\cite{liang2023rich} comprises 18,000 image-text pairs from the Pick a Pic dataset~\cite{Kirstain2023PickaPicAO}. Each image includes human-provided annotations: two heatmaps indicating artifact/implausibility and misalignment, four scores (plausibility, alignment, aesthetic, and overall quality), plus text for misaligned keywords. To ensure photorealism and balance across classes, the PaLI visual question answering model evaluates realism, selecting images from these five classes: \textit{animal}, \textit{human}, \textit{object}, \textit{indoor scene}, and \textit{outdoor scene}. Heatmaps are generated by averaging annotators' key point data related to artifact and misalignment. This dataset supplies intricate annotations to fine-tune scoring models with human feedback. Nonetheless, with only 27 annotators and around 3,000 rater-hours, concerns arise about annotation quality and the reliability of a limited rater pool.

DrawBench is a dataset proposed by Saharia\etal\cite{saharia2022photorealistic}, developed alongside the Imagen model. It comprises a challenging set of 200 prompts designed to evaluate T2I generators across 11 categories, aimed at investigating various abilities such as colors, numbers of objects, spatial relations, text in the scene, unusual interactions between objects, misspellings, rare words, long prompts, and prompts from Reddit, Gary Marcus\etal\cite{marcus2022preliminary}, and DALL-E\cite{ramesh2021zero}. Saharia\etal\cite{saharia2022photorealistic} utilizes DrawBench to compare different T2I models; thus, they present generated images to human raters for quantifying image quality and text-image alignment quality.

PaintSkills proposed by Cho\etal\cite{Cho_2023_ICCV} constitutes a dataset collected specifically to mitigate a statistical bias towards a few common objects. Therefore, Cho\etal generates a dataset carefully controlling three aspects (skills): object recognition, object counting, and spatial relations resulting in 65,535 scene configurations. By uniformly sampling from a set of relations, PaintSkills ensures equally distributed objects and relations. Finally, based on the scene configurations, a 3D simulator is used to render images.

Feng\etal\cite{feng2023trainingfree} proposes two datasets, Attribute Binding Contrast (ABC-6K) and Concept Conjunction 500 (CC-500). The former dataset is derived from MSCOCO, where Feng\etal filters for sentences containing at least two color words, and by switching the position of two color words, they generate additional contrastive sentences, resulting in a total of 6.4K sentences. CC-500 is generated by combining two objects with their attribute descriptions, where each sentence follows the same pattern, e.g. "a red apple and a yellow banana," resulting in 500 sentences.

The Inappropriate Image Prompts (I2P) dataset~\cite{Schramowski_2023_CVPR} targets safe latent diffusion by mitigating the problem of models generating inappropriate images. Therefore, Schramowski\etal collected 4,703 prompts from an online source that distributes real-world human-generated prompts together with SD~\cite{Rombach_2022_CVPR} generated images and corresponding generation parameters. Prompts are filtered based on 26 keywords that correspond to one of seven inappropriateness concepts, e.g. hate, harassment, violence, self-harm, sexual content, shocking images, and illegal activity.

Holistic Evaluation of Text-to-Image Models (HEIM)~\cite{lee2023holistic} is a benchmark dataset that evaluates T2I models based on 12 aspects, e.g. alignment, quality, aesthetics, originality, reasoning, knowledge, bias, toxicity, fairness, robustness, multilingualism, and efficiency. It combines several existing text-image datasets like MSCOCO, DrawBench, PartiPrompts, Winoground, PaintSkills, I2P, etc. to cover the evaluation of each aspect. 

The goal of the SeeTRUE benchmark is to study text-image alignment evaluation. The dataset builds on top of several existing vision-language datasets: COCO Captions~\cite{chen2015microsoft}, SNLI-VE~\cite{xie2019visual}, DrawBench~\cite{saharia2022photorealistic}, EditBench~\cite{Wang_2023_CVPR}, Winoground~\cite{Thrush_2022_CVPR} and Pick a Pic~\cite{Kirstain2023PickaPicAO}. It includes $31,855$ real and synthetic image-text pairs and corresponding human annotations, where each binary annotation indicates alignment or misalignment of text and image.

\section{Open Challenges}\label{sec:challenges}

Within this section, we highlight some of the open challenges that we discovered when reviewing the described T2I quality metrics.

\noindent\textbf{Uncertainty vs alignment quality.} Measuring text-image alignment focuses on relations between objects described by a text. That includes spatial and non-spatial relations between objects and their bound visual attributes. However, these alignment-focused metrics are targeted to sense the presence or absence of certain compositions in image space, but are unable to quantify the quality of such detected components (if present). Quality scores provided by VLM-based metrics are defined on their visio-linguistic capability providing quantitative reasoning in the form of class probabilities for \textit{Yes} or \textit{No} answers to closed questions. However, such a probability score merely indicates the degree of uncertainty rather than actual alignment quality. Future measures should be designed to compute quantities for detected compositions enabling them to rank alignment quality on a component level rather than on the basis of uncertainty scores.

\noindent\textbf{Bags-of-word behavior.} VLMs tend to behave like bags-of-words~\cite{yuksekgonul2022and}, which is a phenomenon that describes a model's insensibility to word order and permutations of object relations, \eg{} the text-image alignment of the sentences "the goldfish is swimming in the aquarium" and "the aquarium is swimming in the goldfish" are scored similarly. Such behavior is caused by the training objective applied to pre-train VLMs. The contrastive pre-training optimizes for image-text retrieval on large datasets, which does not acknowledge compositional information and thus fails to learn unique representations~\cite{yuksekgonul2022and}. A step towards a solution to this problem is hard negative samples~\cite{yuksekgonul2022and,castro2024clove}, where existing prompts are transformed to represent negative compositional semantics by word or relation swapping, and are included in the training set for fine-tuning.

\noindent\textbf{VLM halucination.} Many of the content-based T2I metrics rely on the outputs of VLMs or LLMs that may contain additional details fabricated by a language model rather than actually represented by the image. Additionally, VLMs show limited capability of understanding inputs of multiple images, which may result in low correlation scores on image editing tasks~\cite{ku2023viescore}. VLMs are good at generation task evaluation, but fail at image-to-image evaluation due to high-level feature focus~\cite{ku2023viescore}. Limited context size may reduce the capability of understanding complex text inputs, resulting in discrepancies when mapping an entire image to sparse text tokens~\cite{urbanek2023picture}. 

\noindent\textbf{Dataset availability.} Existing T2I datasets~\cite{lin2014microsoft,young2014flickr30k,mao2016training,sharma2018conceptual,Changpinyo_2021_CVPR} mainly originate from various online sources, where image-text pairs are collected by applying heuristics to filter the data, thereby often trading quality for quantity. Otherwise, high-quality image descriptions need to be crowdsourced by human annotators, which is time-consuming and costly. With increasing focus towards the evaluation of visio-linguistic compositionality, the necessity of compositional datasets intensifies~\cite{Thrush_2022_CVPR,Ma_2023_CVPR,hsieh2023sugarcrepe,ray2024cola}. While the evaluation on such complex datasets fosters the development of compositional metrics, the active research in this field seems to stick to a limited set of four compositional aspects: object accuracy, spatial relations, non-spatial relations, and attribute binding. However, we consider this to be a subset of a greater set which is yet to be explored; thus, in the work of Dehouche~\cite{dehouche2023s} they apply GPT-3~\cite{openai2023gpt} to explore a set of 20 topics: \eg{} medium, technique, genre, mood, tone, lighting, artistic reference, which are derived from human-generated prompts taken from Lexica\footnote{\scriptsize\url{https://lexica.art/}}. 

Further, benchmarking image generation is lacking comparability due to evaluation on individually proposed datasets providing insights on specific topics. Although there are widely adopted compositional datasets~\cite{Thrush_2022_CVPR}, the size of such datasets limits the assumptions that can be made regarding generalizability. However, creating a comprehensive benchmark for compositionality evaluation should be targeted in the near future.

\section{Guidelines}\label{sec:guidelines}

In the following, we provide guidelines for evaluating T2I generation models based on our findings surveying the literature. These guidelines are formulated with the goal to help researchers and practitioners to make more informed choices about which metrics and benchmark datasets to use when working with T2I generation.


\noindent\textbf{Select metrics based on relevant characteristics.} Benchmarking T2I generation involves measuring general image quality and compositional quality (cf. \Cref{sec:taxonomy}). However, what defines the image quality might depend on the target application. For instance, in the domain of artistic image generation (\eg{} comics, anime, mangas, and paintings) an image has to reflect certain art styles, drawing characteristics, shapes, and colors. However, images do not need to be photo-realistic and naturalistic. In order to capture and measure such a large variety of abstract concepts, there exist many visual quality metrics, see \Cref{tab:text-image-alignment-metrics}. Each of these metrics is equipped with unique reasoning capabilities, such as aesthetic and human preference prediction, perceptual artifact localization, object recognition, object counting, spatial relations, object attribute recognition, and many more. Hence, reasoning skills for evaluation techniques need to be selected carefully and the calibration of their priority is crucial. Considering the use case of generating synthetic images for pre-training object detection networks of real images, one would need to ensure that the image generator produces correct visual representations of described objects. This necessitates metrics with strong object recognition and object counting capabilities. As shown in \Cref{sec:metrics}, the current state-of-the-art does not include a general purpose metric satisfying a comprehensive evaluation of T2I generation. We provide a classification of metrics and their capabilities, which can be used to make informed decisions about which metric to use in a specific application context.

\noindent\textbf{Select appropriate evaluation prompts.} The underlying text prompts are fundamental for evaluating T2I generation, as they form the input to the image generator. Equally important to selecting the right metrics is ensuring that evaluation prompts include rich descriptions that cover a broad set of visual concepts. Otherwise, there is no way to obtain comprehensive benchmark results. In Section~\ref{sec:datasets}, we provide an overview of the state-of-the-art datasets containing image-text pairs with different levels of complexity. Textual descriptions that originate from image captioning datasets usually lack the range of visual concepts needed for the evaluation of T2I generation. Using prompts that do not cover the visual depictions to be measured can help outperform other methods but render test results meaningless. Therefore, the collection of evaluation prompts needs to represent authenticity, complexity, compositionality, and representativity of textual descriptions with respect to the target application. 

\noindent\textbf{Normalize prompts.} The most recent diffusion-based image generation models~\cite{yang2023diffusion} can be used to synthesize realistic-looking images of impressive quality. The data these models were trained on may be subject to language bias, which results in a biased image generator, \eg{} specific sentence formats, such as the absence of grammatical structure, certain keyword constellations, or artist names that are known only by some models. In order to mitigate such bias, it can help to normalize evaluation prompts by adopting the strong rephrasing, summarization, and completion capabilities of modern LLMs. In particular, LLMs can be used to transform prompts to natural language, complete sentences, and remove keywords. On top of this, further normalization protocols may be applied. For some applications, it may be beneficial to normalize prompt length since some text encoder networks have limited token vector lengths. Hypernymization, where a word is replaced by its hypernym (\ie{} another word that describes it in a more general way, \eg{} daisy and rose would be replaced by flower), is a method to semantically normalize prompts~\cite{baryshnikov2023hypernymy}. However, this may lower the variety of evaluation prompts. Furthermore, the representation of numbers and dates can be brought into a consistent format, \eg{} \textit{Two dogs are playing with a ball.} and \textit{2 dogs are playing with a ball}.

\noindent\textbf{Set model parameters.} As diffusion-based image generation is sensitive to the selected seed for the initial noise sampling during early diffusion steps, it is crucial to fix such seeds to guarantee reproducibility. Further, some benchmarks compare image generators that share an identical training protocol and have only small architectural differences or vice versa. Utilizing identical seeds clarifies the contribution of these changes. The image resolution used during training can have a strong influence on image quality. Thus, it should be configured appropriately and consistently throughout all evaluated methods. This extends to the sampling method, sampling steps, and guidance parameters. When the image generation pipelines are properly configured for each prompt in the evaluation set, a fixed number of $N$ images is generated, where $N$ is equal to the number of model parameter configurations. Higher numbers for $N$ provide increasingly robust performance results in exchange for computational costs.

\section{Conclusion}\label{sec:conclusion}
This survey provides an overview of the current state of the art in evaluation metrics for T2I generation. First, we introduced our taxonomy to categorize quality measures based on the data they evaluate (images alone vs. text and images), their scope (distribution of images vs. single images), their operating data structure (embeddings vs. content), and what they measure (general quality vs. compositional quality). Many widely adopted T2I metrics lack the ability to assess the alignment between text and image, and thus can omit important details during evaluation. To develop the proposed taxonomy, we collected, reviewed, and compared both established and emerging evaluation metrics, acknowledging the trend toward compositional quality metrics, which are sensitive to the prompt definition and can detect and judge the model’s alignment quality between image and text. Furthermore, we have reviewed existing text-image datasets. Many of these datasets are specifically tailored to benchmark visio-linguistic compositionality but possess an insufficient amount of data for comprehensive T2I generation evaluation, thus lacking compositional evaluation features. Based on our observations made, we further discussed open challenges of existing evaluation methods. The importance of the quality metrics reviewed extends even beyond the T2I domain. In fact, it is fundamental for application areas such as text-to-video~\cite{ho2022video,ho2022imagen,zhang2023spot}, where multiple frames are generated for a single text prompt, and text-to-3D, where image-based NeRF approaches and diffusion models produce 3D representations for textual scene descriptions~\cite{Lin_2023_CVPR,Metzer_2023_CVPR,Chen_2023_ICCV,Wang_2023_CVPR,wang2023prolificdreamer}. In these broader contexts, robust and reliable metrics are essential to assess alignment and compositionality across more complex or temporal data. Finally, based on all our findings, we provide guidelines for the development of comparable and meaningful evaluation protocols. These guidelines will enable consistent quality assessment and, thus, representative T2I generation evaluation.

\ifCLASSOPTIONcaptionsoff
  \newpage
\fi



%
\bibliography{main}
\bibliographystyle{IEEEtran}

%

\vspace{-40pt}
\begin{IEEEbiography}[{\includegraphics[width=1in,height=1.25in,clip,keepaspectratio]{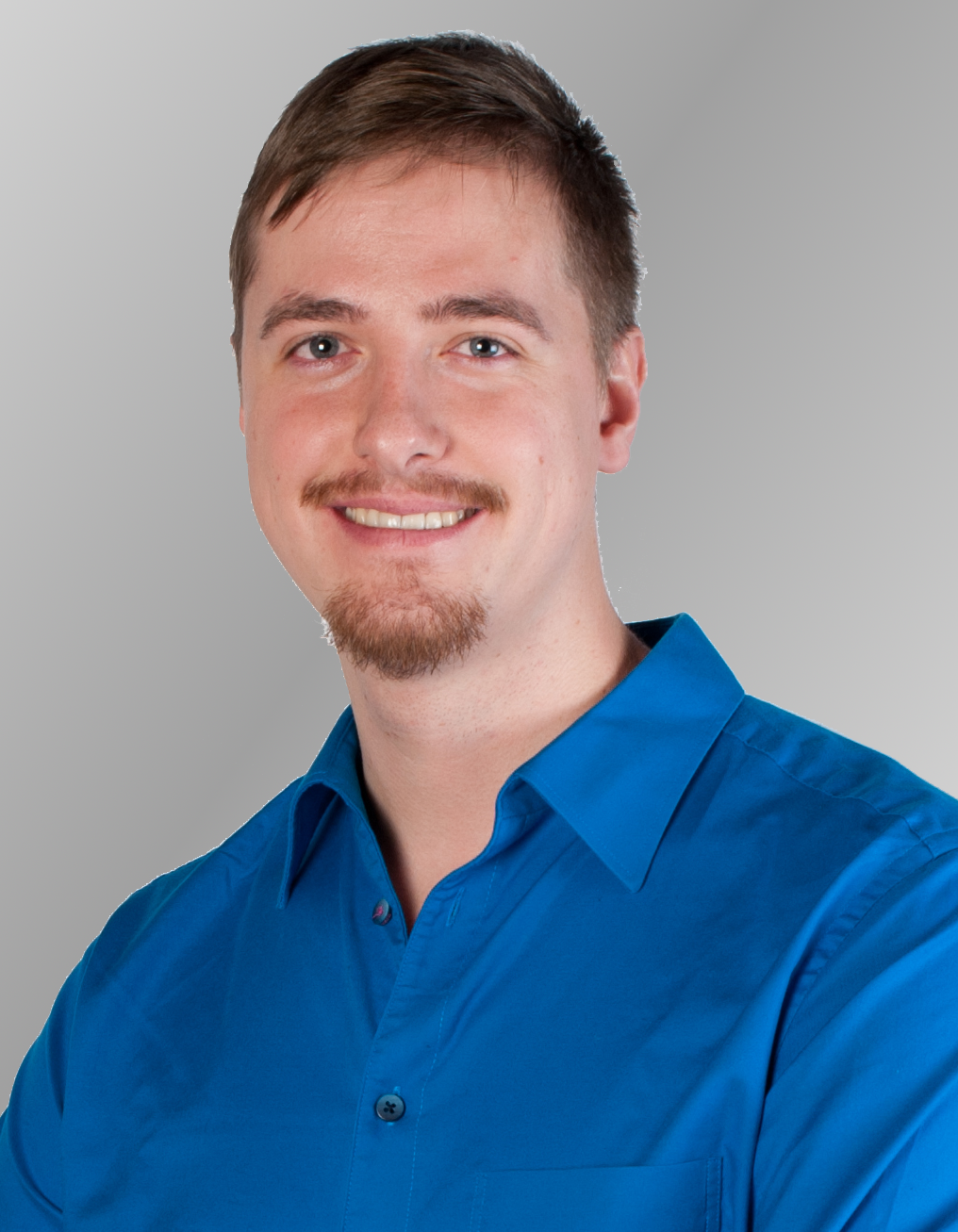}}]{Sebastian~Hartwig}
is a Ph.D. student at Ulm University, Germany, where he previously received his B.Sc. and M.Sc. degrees in computer science. He completed his master’s degree at the Institute of Media Informatics in 2017 before joining the research group Visual Computing. His research focus on human perception based machine learning for computer graphics and visualization application, as well as machine scene understanding like depth and layout estimation.
\end{IEEEbiography}

\vspace{-35pt}

\begin{IEEEbiography}[{\includegraphics[width=1in,height=1.25in,clip,keepaspectratio]{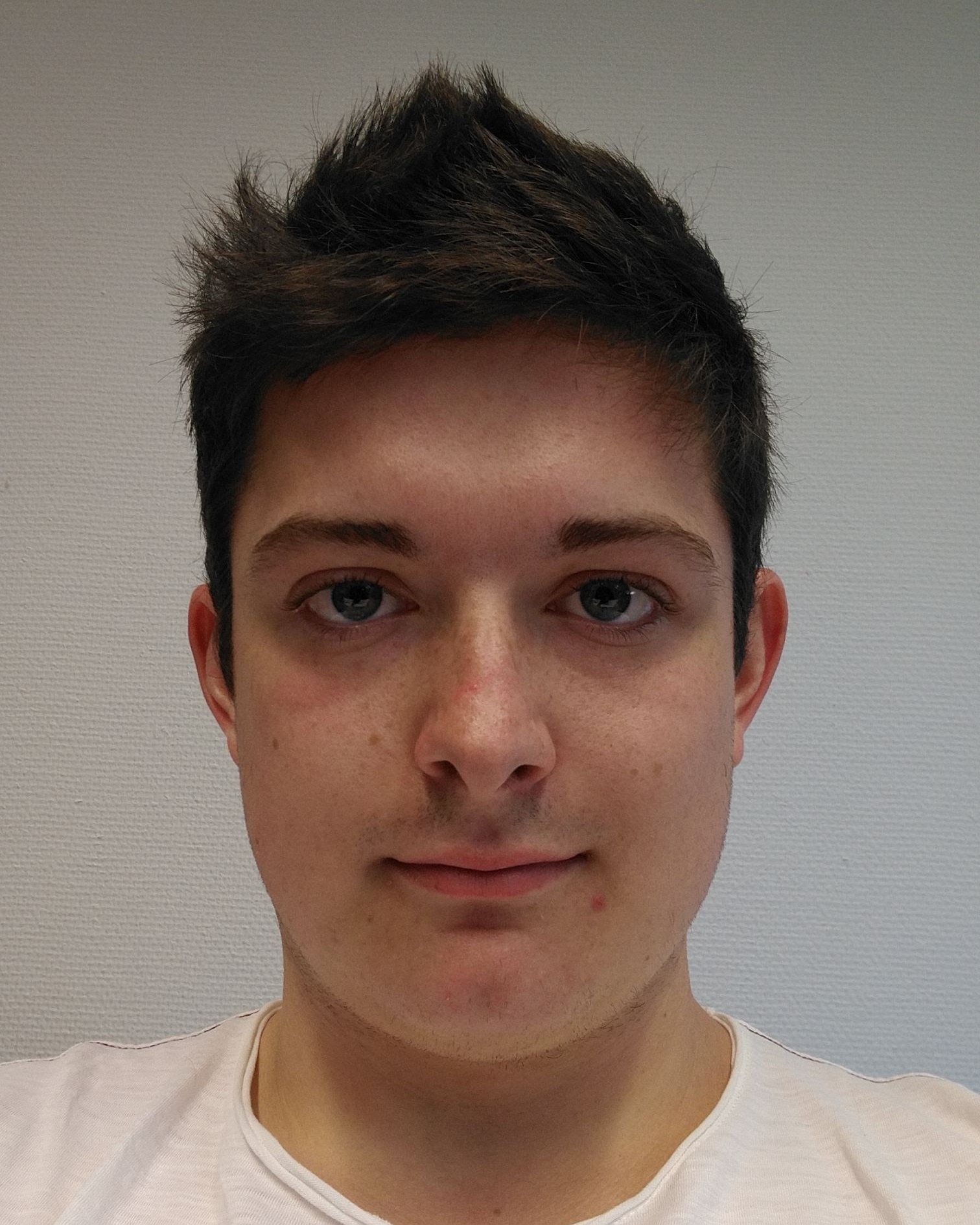}}]{Dominik~Engel}
is a Ph.D. student at Ulm University, Germany, where he previously received his B.Sc. and M.Sc. degrees in computer science. In 2018, he joined the Visual Computing research group. His research focuses on deep learning in visualization and computer graphics, differentiable and neural rendering.
\end{IEEEbiography}

\vspace{-40pt}

\begin{IEEEbiography}[{\includegraphics[width=1in,height=1.25in,clip,keepaspectratio]{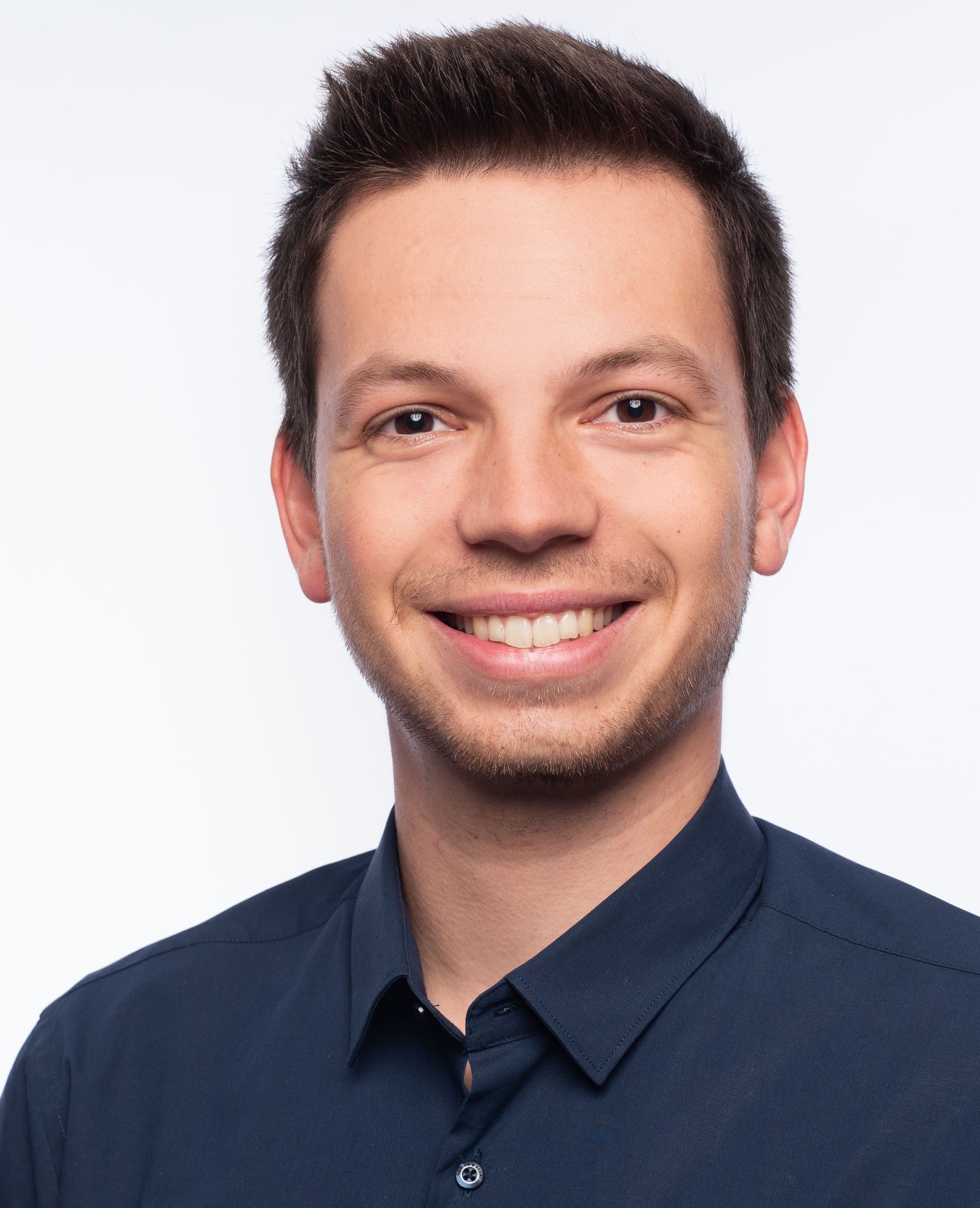}}]{Leon~Sick}
is a Ph.D. student at Ulm University and part of the Visual Computing Group. Before starting his Ph.D., he obtained his B.A. in International Business Administration from Aalen University of Applied Sciences and his M.Sc. in Business Information Technology from Konstanz University of Applied Sciences. His research is focused on self-supervised pre-training and unsupervised segmentation on 2D images.
\end{IEEEbiography}

\vspace{-40pt}

\begin{IEEEbiography}[{\includegraphics[width=1in,height=1.25in,clip,keepaspectratio]{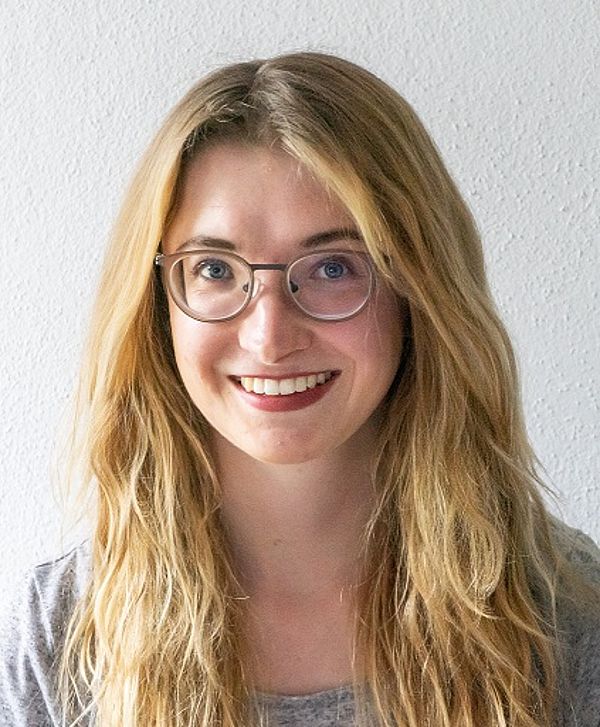}}]{Hannah~Kniesel}
is a Ph.D. student at Ulm University, Germany. She finished her M.Sc. in 2022 with a focus on the reconstruction of bio-medical data. Prior to that, she completed her Bachelor's degree in 2019, also at the University of Ulm. She joined the Visual Computing Research Group in April 2020.
\end{IEEEbiography}

\vspace{-40pt}

\begin{IEEEbiography}[{\includegraphics[width=1in,height=1.25in,clip,keepaspectratio]{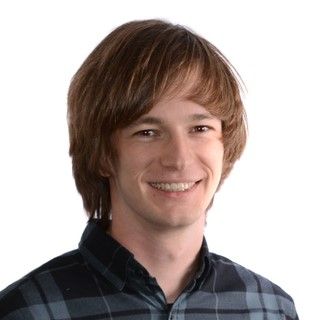}}]{Tristan~Payer}
has joined the Visual Computing Group in November 2020 as a research associate. He received his master's degree in 2020 from Radboud University Nijmegen in the field of Artificial Intelligence. His focus in the master's program was on Natural Language Processing, Medical Image Analysis, and Deep Learning. Previously, he completed his bachelor's degree at Radboud University Nijmegen in 2018.
\end{IEEEbiography}

\vspace{-40pt}

\begin{IEEEbiography}[{\includegraphics[width=1in,height=1.25in,clip,keepaspectratio]{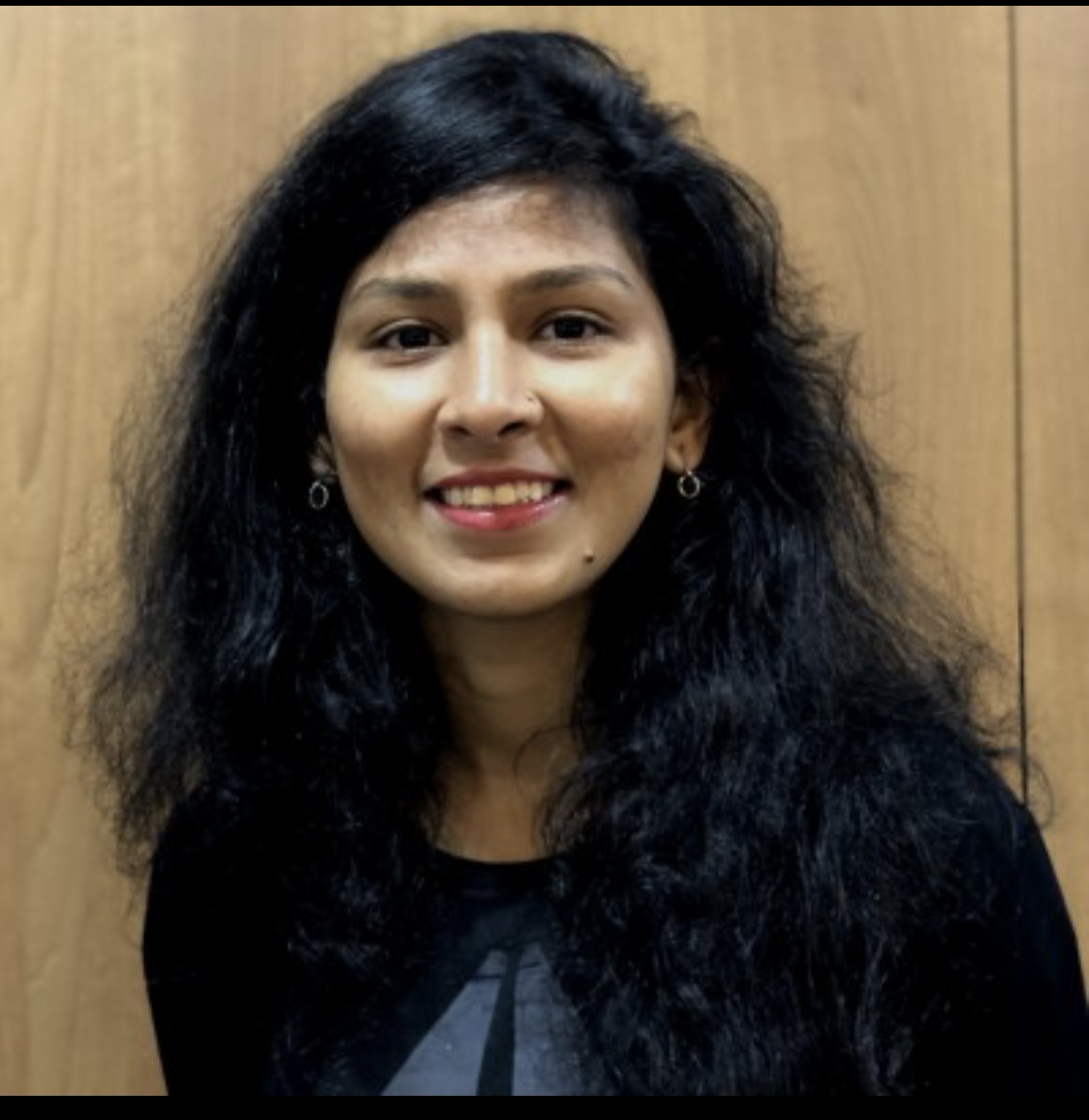}}]{Poonam Poonam}
has completed her Master in Cognitive Systems at Ulm University in January 2023 before joining the research group Visual Computing. In her Master thesis, she focused on contrastive and continual learning.
\end{IEEEbiography}

\vspace{-40pt}

\begin{IEEEbiography}[{\includegraphics[width=1in,height=1.25in,clip,keepaspectratio]{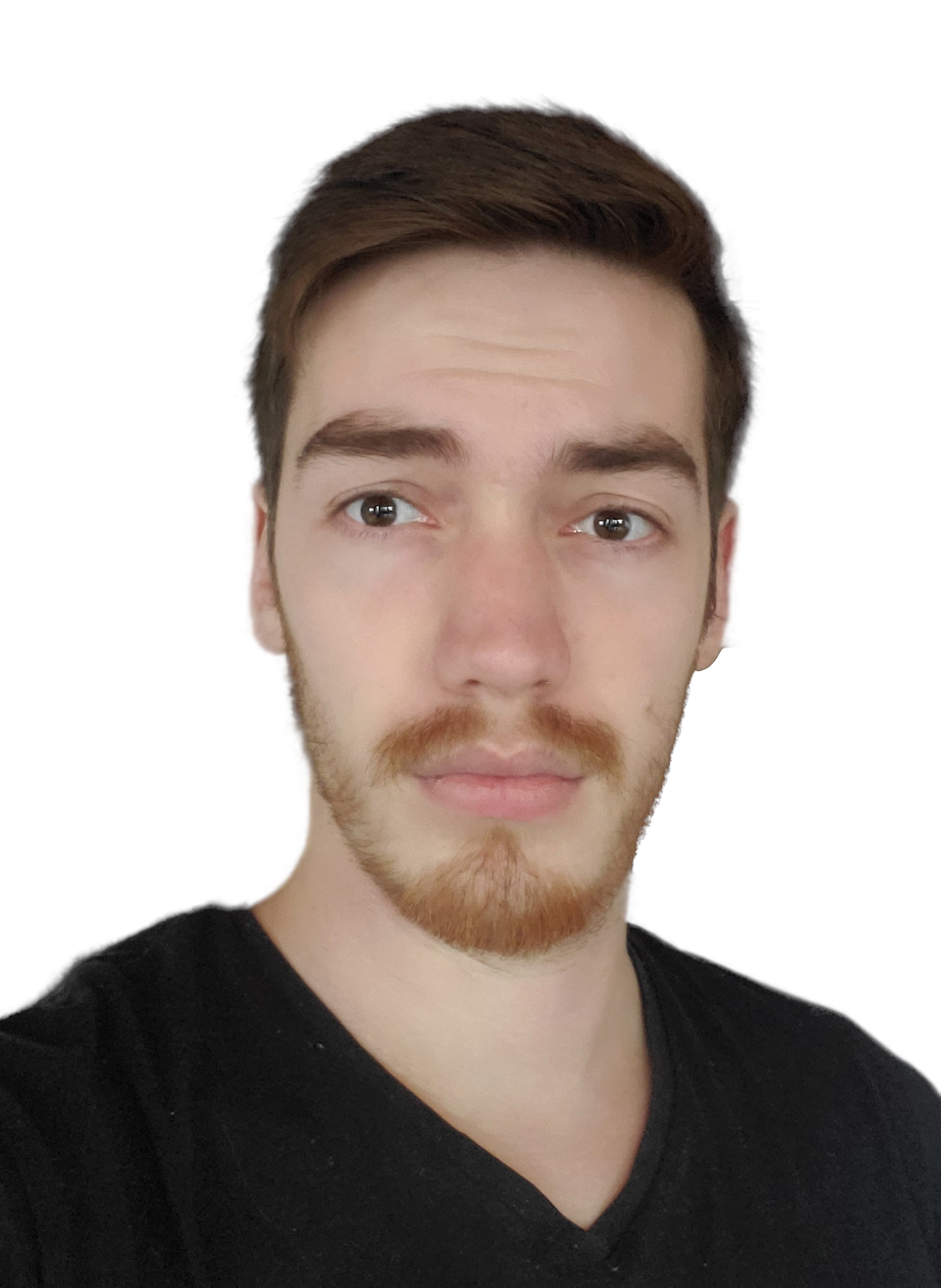}}]{Michael Glöckler}
is currently pursuing a Master's degree in Media Informatics at Ulm University, having previously earned his Bachelor's degree in the same field from Ulm University in 2022. Collaborating on projects with the Visual Computing Group, exploring areas such as natural language processing,
deep learning, and T2I generation.
\end{IEEEbiography}

\vspace{-40pt}

\begin{IEEEbiography}[{\includegraphics[width=1in,height=1.25in,clip,keepaspectratio]{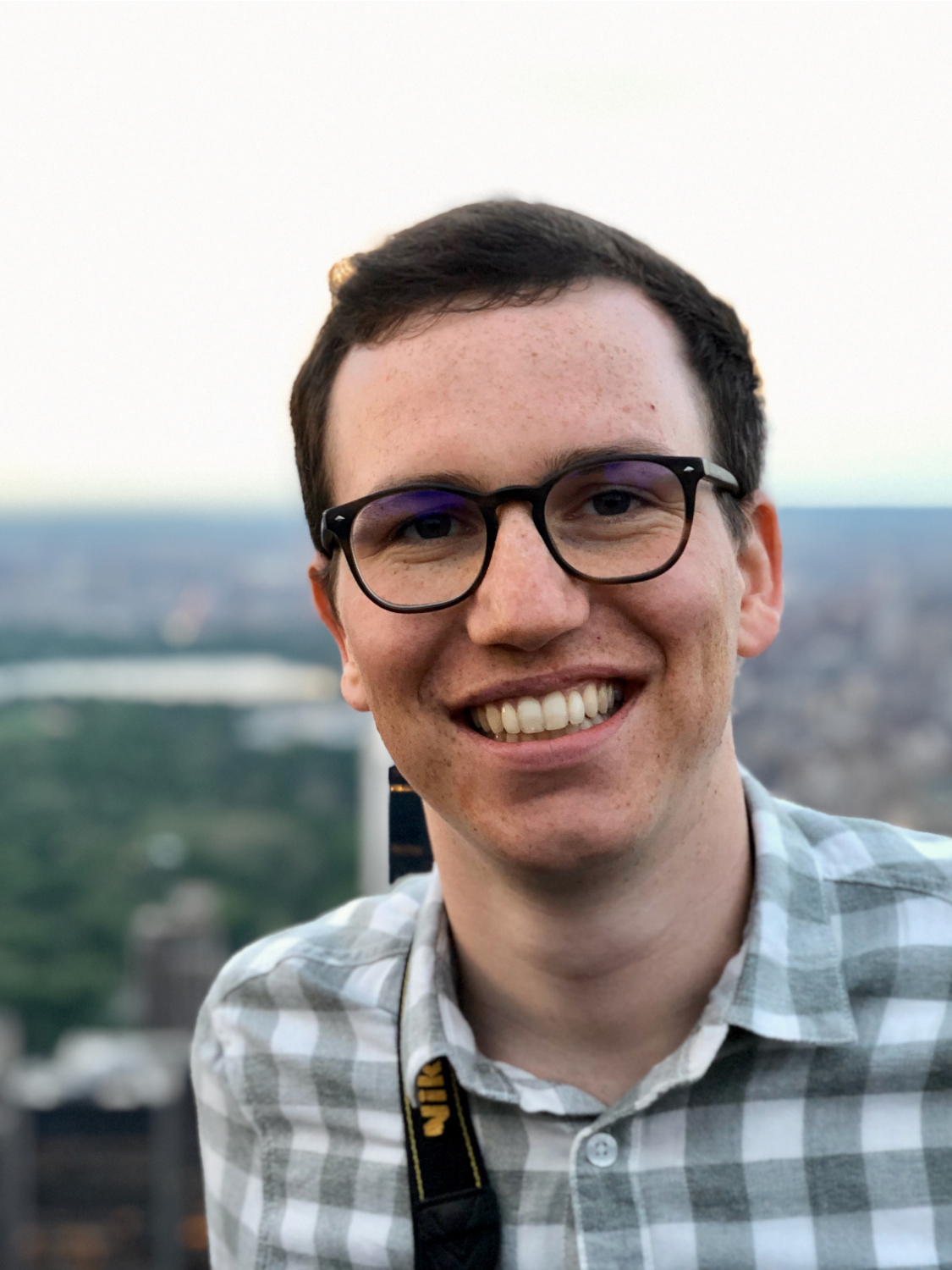}}]{Alex Bäuerle}
  received his Ph.D. in computer science from Ulm University in 2022; afterwards, he worked as a researcher at Carnegie Mellon University. Currently, Alex is a Founding Member of Technical Staff at Axiom, where he works on ML for liver toxicity prediction. His current research interests are at the intersection of human-computer interaction and artificial intelligence.
\end{IEEEbiography}

\vspace{-40pt}

\begin{IEEEbiography}[{\includegraphics[width=1in,height=1.25in,clip,keepaspectratio]{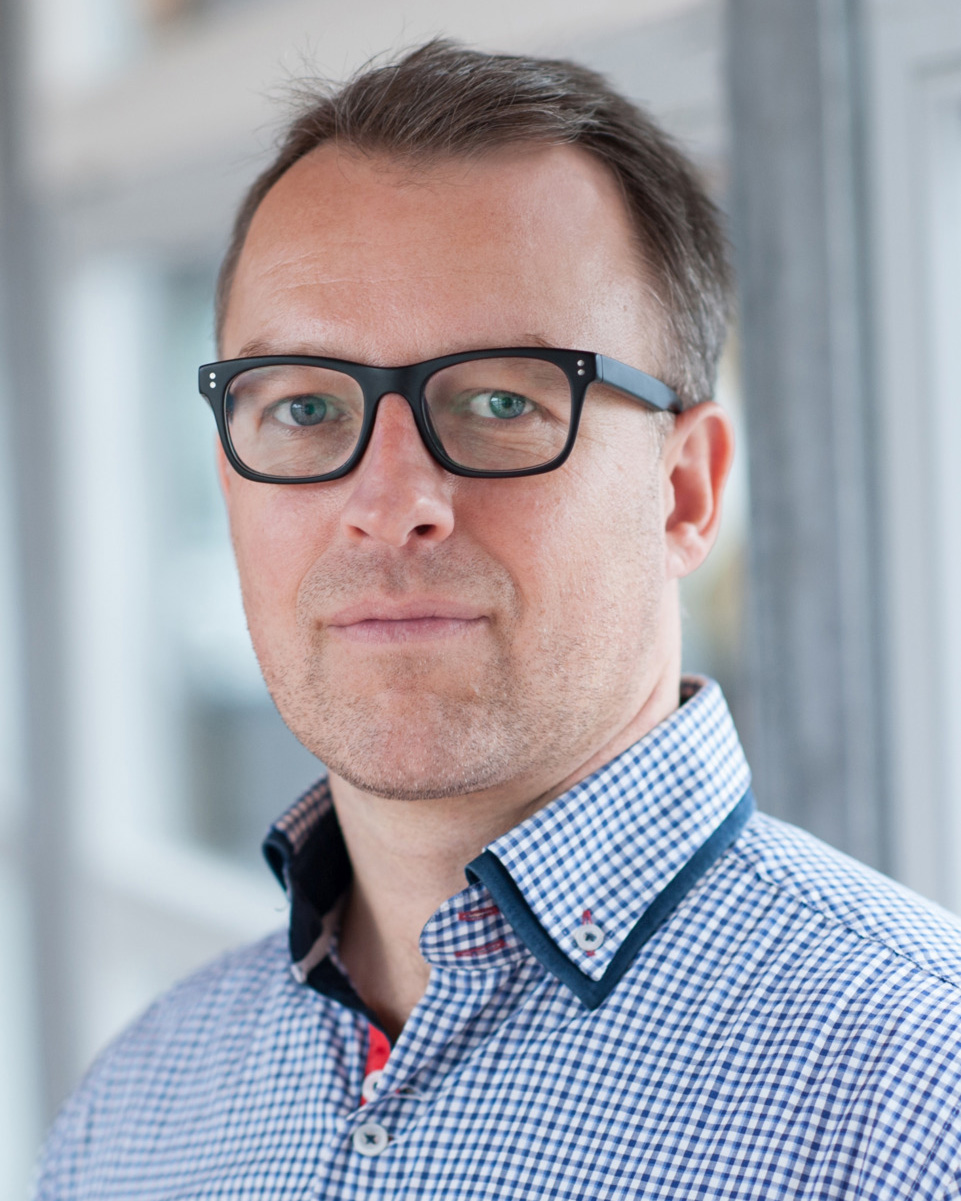}}]{Timo~Ropinski}
is a professor at Ulm University, heading the Visual Computing Group. Before moving to Ulm, he was Professor in Interactive Visualization at Linköping University, heading the Scientific Visualization Group. He received his Ph.D. in computer science in 2004 from the University of Münster, where he also completed his Habilitation in 2009. Currently, Timo serves as chair of the EG VCBM Steering Committee, and as an editorial board member of IEEE TVCG.
\end{IEEEbiography}








\section*{Supplementary material}

# Supplementary Material
# A Survey on Quality Metrics
# for Text-to-Image Generation

Sebastian Hartwig⬤, Dominik Engel⬤, Leon Sick, Hannah Kniesel⬤, Tristan Payer⬤,
Poonam Poonam⬤, Michael Glöckler⬤, Alex Bäuerle⬤, Timo Ropinski⬤

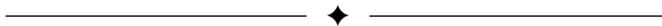

## 1 OPTIMIZATION

While T2I quality metrics are developed to judge the quality of single T2I samples or complete T2I models, other techniques can also be used to optimize the generated output. In this section, we will review some of these techniques, whereby we differentiate between those which require a training process (see Section 1.1), and those which can be used without additional training (see Section 1.2).

### 1.1 Fine-tuning Image Generators

The StyleT2I framework [1], which utilizes a CLIP-guided contrastive loss, a semantic matching loss, and a spatial constraint to refine attribute manipulation within intended spatial regions, trains a StyleGAN [2] model to increase compositional accuracy. This approach ensures more disentangled latent representations that can be decoded into a high-fidelity image aligned with the input text. In order to find optimal latent codes, a Text-to-Direction module is employed to predict the sentence direction that is aligned with the input text, which is trained using a CLIP-guided contrastive loss. To enhance attribute alignment, a *Attribute-to-Direction module* is optimized by the semantic matching loss that seeks to identify attribute directions of latent codes. To mitigate the issue of changing multiple regions during attribute alignment, a spatial constraint avoids spatial variations outside of a pseudo-ground-truth mask generated by a segmentation model.

In the work of Dong et al. [3], called RAFT, a Stable Diffusion model is fine-tuned on high-quality samples ranked by a reward model. In a decoupled data generation process, high-quality training data is sampled from the generator by discarding low-quality data points, which is done by utilizing a reward model to filter out those that exhibit undesired behavior. In their experiments, they adopt CLIPScore and LAION Aesthetic Predictor as reward models; however, fine-tuning on training data produced by the generator hinders the model from overcoming problems within distribution towards high-quality compositionality, which remains

concealed outside the generator's distribution. After training a reward model utilizing their proposed dataset, see **??**, Liang et al. [4] demonstrate fine-tuning a Muse image generator and compare it to the pre-trained Muse version based on 100 generated test prompts. In their experiments, they use Muse to generate a set of $100,512$ images for $12,564$ generated prompts, then they apply their reward scores to filter out images below a certain threshold, and fine-tune Muse. Finally, they quantify the gain from Muse fine-tuning by conducting a study, where they present the two images side-by-side originating from the baseline and fine-tuned model. They were able to show that the fine-tuned model produces significantly fewer artifacts/implausibilities than the original Muse. A similar approach by Lee et al. [5] collects binary decisions from human annotators for a large set of generated images to train a reward model. Thus, they are able to fine-tune Stable Diffusion via reward-weighted likelihood maximization to better align it to human feedback using 27K image-text pairs. In addition, AlignProp [6] is following the idea of utilizing a reward model to supervise the fine-tuning of Stable Diffusion with human feedback.

After such recent developments of reward models opening a way towards incorporating human feedback into the diffusion process, further approaches adopted reward models, applying supervision during reinforcement learning. Thus, Fan et al. [7] propose *DPOK*, diffusion policy optimization with KL regularization, which utilizes KL regularization to stabilize RL fine-tuning and aligning T2I. Unlike traditional supervised fine-tuning, which often degrades image quality, RL fine-tuning with DPOK optimizes ImageReward [8] a feedback-trained reward model online, leading to better alignment between text prompts and generated images while maintaining high image fidelity. The paper demonstrates that DPOK outperforms supervised fine-tuning methods in experiments, showcasing its effectiveness in enhancing T2I diffusion models. However, their work studies KL-regularization and primarily focuses on training a different diffusion model for each prompt. Denoising diffusion policy optimization (DDPO), proposed by Black et al. [9] broadens the approach by training using multiple prompts and showcasing the model's ability to

generalize to unseen prompts. The adaptability of DDPO to various reward functions, including those derived from vision-language models, marks its advance in enhancing prompt-image alignment beyond the scope of human feedback optimization.

## 1.2 Training free Optimization

In their work, Liu et al. [10] introduce an approach called Composable Diffusion Models, targeting the challenge of accurately aligning compositional text prompts to their image representation. Their method proposes a structured strategy where an image is generated through the composition of a set of diffusion models, each modeling different visual concepts. By treating diffusion models as energy-based models, they enable the explicit combination of data distributions defined by these models. This approach assumes that the visual concepts are conditionally independent given the image. Sampling from the resultant distribution involves using a composed score function that integrates the contributions of each concept to the denoising process, allowing for the generation of images that faithfully represent the composed concepts. This allows for the generation of scenes that are more complex than those encountered during training, effectively combining object relations and attributes accurately. Feng et al. [11] introduces *Structured Diffusion Guidance*, a method aimed at improving compositional T2I generation through the use of scene graphs, which are derived from the prompt. Instead of computing a text embedding for a single sequence, this method extracts noun phrases from the prompt corresponding to visual concepts and entities, and encodes such noun phrases separately to achieve region-wise semantic guidance. Finally, for each cross-attention map, the average of all noun phrase activations denotes the corresponding output. The method *Attend and Excite* by Chefer et al. [12] introduces a training-free approach called *Generative Semantic Nursing* (GSN), which utilizes cross-attention maps from a Stable Diffusion model [13] to enhance the representation of objects and their attributes in images by maximizing attention values at subject regions during inference. This technique improves the incorporation of subject tokens into the latent representation through a loss objective that ensures high activations for at least one patch per subject token. However, the authors note that the timing of this optimization is critical, as object spatial locations become fixed in the final denoising steps, and they advocate for gradual latent refinement to avoid image degradation. While GSN shows promise for simple prompts, its effectiveness diminishes with increased complexity, such as multiple entities with bound attributes. Building on this, Li et al. 's follow-up work, *Divide and Bind* [14], employs total variation maximization to enhance local changes in attention maps, promoting the emergence of diverse object regions and enabling competing objects during generation. They introduce two new objectives: one for attending to object tokens and another for attribute binding regularization, utilizing a finite differences approximation of total variation to create activation patterns that segment the image. Additionally, they normalize overlapping object and attribute tokens in the attention maps and maximize their symmetric similarity by minimizing the

Jensen-Shannon divergence [15]. The metric proposed by Singh et al. [16], see ?? for further details, is able to detect parts of the image that are not aligned with the prompt, and when passing the relative importance of such parts to a modified reverse diffusion process, it becomes possible to improve T2I alignment. In their experiment, they can show that by incorporating such relative importance in the form of weighting factors through a combination of prompt weighting and cross-attention guidance (*Attend and Excite*), they can optimize T2I alignment. Layout control with cross-attention guidance is achieved by Chen et al. [17], by introducing two techniques to incorporate object bounding boxes to steer the diffusion process optimizing for spatial relation alignment. This method enables control of image generation by providing object bounding boxes to encourage the diffusion model to generate corresponding objects within the bounding box region of the image. The work proposes two ways of layout guidance: forward and backward guidance. The former method applies a smoothed window function to the cross-attention maps, which increases activation for corresponding object tokens inside the bounding box. The second method, backward guidance, applies an energy-loss function which is computed via back-propagation to update the latent vector and therefore indirectly alter the cross-attention maps. During image generation, they alternate between denoising steps and gradient updates. While this technique improves overall layout and provides control over spatial relations, problems of standard diffusion approaches for object attribution remain and the question of manual bounding box placement needs to be considered.

## 2 Human Preference Metrics

In the following experiments, we investigate human preferences regarding text prompt generation and corresponding diffusion-based image generation. Consequently, our first goal is to inspect human-provided text prompts, investigating preferred styles, concepts, topics, sceneries, etc., which we retrieve in a data-driven way from a large corpus of human-provided data.

### 2.1 Dataset Acquisition

In order to collect a large-scale dataset consisting of human-generated prompts along with generated images, we downloaded messages from the official Midjourney Discord server. Each message contains information about a user-generated text prompt, the corresponding generated image, the author's name, as well as links to preceding and succeeding user interactions. In this way, we were able to collect $8,290,132$ text-image pairs.

### 2.2 Data-driven Prompt Categorization

To gain initial insights into the dataset containing various prompts, we examined a subset and discovered that users often use the same prompt multiple times with only minor modifications to generate an image. This observation led us to infer that users were dissatisfied with the initial image generated, prompting them to modify the prompt in hopes of achieving a more pleasing outcome. To understand what users added to the prompts to enhance image generation,

we conducted a detailed analysis of these modified prompts. Establishing relationships between these prompts requires a preliminary comparison of their similarity. To accomplish this, we examined all prompts in the dataset and calculated CLIP embeddings for each individual prompt. With these CLIP embeddings, we were able to compare prompts using cosine similarity between the embeddings and connect prompts with high similarity in a graph structure. This enables us to understand user interactions with multiple prompts and their corresponding images produced by Midjourney. Furthermore, we assume that the most recent prompt, in terms of time, was considered the best variant by the user, as it seemingly met their satisfaction, leading them to cease modifications of the prompt. Subsequently, we examined all final prompts from the prompt graphs and created a dictionary to store the occurrences of each individual word.

| Category | Top Words |
|---|---|
| realistic | **"realistic"** (6820), "real" (674) |
| logo | **"logo"** (5789) |
| photo | "lighting" (5056), "photography" (1940), **"photo"** (1778), "photorealistic" (1248), "reflections" (1137) |
| Art | **"art"** (2312) |
| cartoon | **"cartoon"** (2064) |
| anime | **"anime"** (1699), "manga" (185) |
| cyberpunk | **"cyberpunk"** (1207), "futuristic" (1061), "cyber" (182) |
| portrait | **"portrait"** (1135), "eyes" (2178) |
| simple | **"simple"** (1065), "minimalist" (795) |
| illustration | **"illustration"** (995), "painting" (970) |
| landscape | **"landscape"** (593), "mountains" (389), "sunset" (673) |

TABLE 1
Categories and Top Words occurred in different prompts

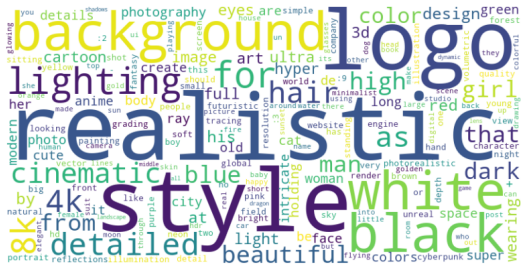

Fig. 1. This wordcloud visualize the occurrences of different words in the final prompts

Then, we proceeded to analyze the most frequently occurring words (cf. Figure 1) and defined categories for the prompts. However, we exclude common words such as "a", "and" and "the" as well as parameter keywords specific to the Midjourney Engine, since they do not contribute to differentiating between categories. The top five most frequent words included general terms such as "style" (6096 occurrences), "background" (5179 occurrences), and "white" (5137 occurrences), as well as category-specific terms like "realistic" (6820 occurrences) and "logo" (5789 occurrences). Through an examination of these frequently occurring words, we were able to define 11 categories, see Table 1. For each category, we then selected a keyword that epitomizes the category. These keywords were subsequently used to ascertain the group to which a prompt belongs.

## 2.3  Inferring Human Preferences

In Section 3 of the main paper, several generated image metrics are reviewed that are based on fine-tuning to increase alignment with human preferences (e.g. ImageReward, Human Preference Score v2, and Aesthetic Predictor). In this experiment, we aim to infer image quality scores, enabling us to investigate the differences between such metrics. As a baseline, we include metrics learned via representation learning, e.g., CLIPScore, BLIP, and BLIP2. In this experiment, we applied the six metrics to all text-image pairs collected in Section 2.1 enabling us to rank

pairs for each metric separately. However, since these scores use different scales for their values; for instance, CLIPScore returns values around 2.5, while the aesthetic score ranges between 0 and 10, comparing the scores is not straightforward. Therefore, we normalized all scores so that all values lie between 0 and 1. In Figure 2, we show box plots of the distribution of each image quality score. It is notable that BLIP mostly contains very high scores, with a median above 0.9 and a very low standard deviation, whereas the other scores demonstrate a more balanced distribution.

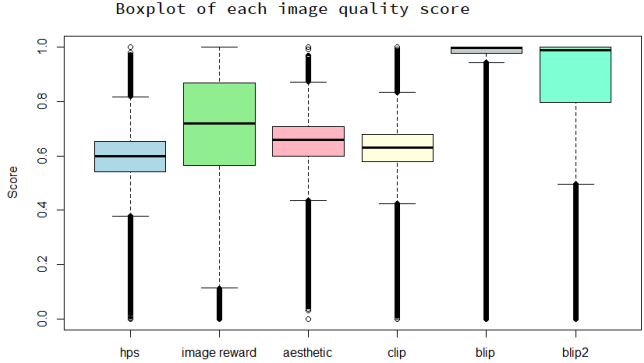

Fig. 2. Box plot visualization of the value ranges for each of the normalized image quality scores.

After taking a closer look at the value ranges for different scores, this section investigates the visual appearance of images ranked by the scores, respectively. Therefore, we attempted to identify the best and worst scored image for each category and quality metric. First, we investigate an overall quality by combining the six metrics, Figure 3 displays the images for each category based on the sum of the six image quality scores. As observed across all categories, the best image exhibits superior characteristics, featuring clear and identifiable features, whether they be animals, humans, or other depicted elements. In contrast, the worst image within each category tends to be harder to recognize, often resulting in unclear visual content or even areas of a single color. This comparison provides valuable insights into the varying quality levels across different categories, highlighting the effectiveness of the selected quality metrics

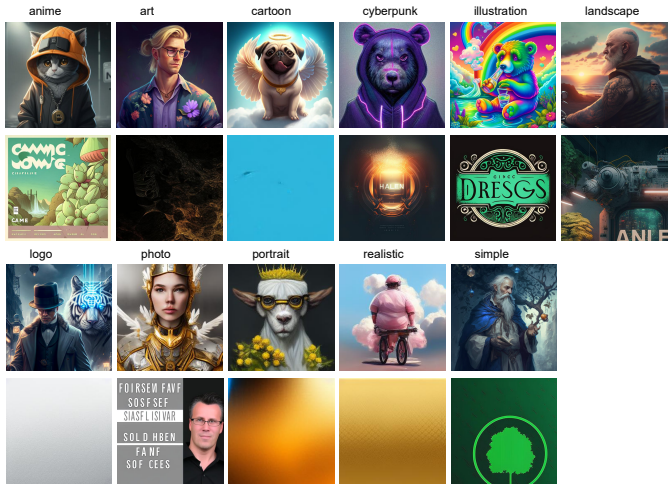

Fig. 3. For each prompt category, we display the corresponding image with the highest (top row) and lowest (bottom row) sum of six image quality scores

in assessing and distinguishing image quality.

To gain more detailed insights, we also examined the highest and lowest-rated images for each individual image quality score. When reviewing the images, as shown in Figure 4, a noticeable difference between the worst and best images is evident, although the contrast is not as strong as it is when considering the sum of all scores, as we did in Figure 3. In some images, the depicted scenes can still be easily identified, which may not always be possible with the lowest-rated images when considering the sum of all scores. Another observation is that CLIP, BLIP, and BLIP2 tend to value a match between the prompt and the image more than the actual aesthetics of the image. The lowest-rated score for an image of these metrics does not correlate with poor visual quality. It seems to receive a lower rating because it doesn't match well with the given prompt. This observation leads to the conclusion that representation learning-based image metrics do not reflect human preferences well.

### 2.4 Influence of Prompt Length

In this experiment, we delve into the correlation between the length of prompts and their evaluation performance across various categories. By analyzing the prompts associated with both the lowest-rated and highest-rated images in each category, as detailed in Table 2, a clear trend emerges: prompts for the lowest-rated images tend to be significantly longer than those for the highest-rated images. This observation suggests that prompt length may inversely affect image evaluation outcomes, potentially due to factors such as clarity, focus, and user engagement. Further analysis reveals patterns in the compositional elements and thematic content of the highest-rated prompts, indicating that optimal prompt length might also depend on the specific context and requirements of each category. Furthermore, we observe additional keywords that remained present after our initial filtering out of command parameters specific to Midjourney. We assume that these characteristics of the prompts introduce a negative bias, resulting in poor text-image alignment assessment by the adopted metrics. This issue should be investigated in future research.

To investigate the potential correlation between prompt length and the combined image quality score, we conducted a detailed examination of the prompts. It was observed that the maximum prompt length is 1800 characters, likely a limitation imposed by MidJourney at the time. Figure 5 illustrates the impact of prompt length on the combined image quality score. It was found that prompts shorter than 200 characters tend to achieve the highest scores, and as prompt length increases, the scores generally decline. However, the correlation between prompt length and image quality scores is not particularly strong, with a coefficient of approximately -0.07. Despite this, the median score remains consistently close to four across all lengths of prompts.

| Best Image | Best Image Prompt | | Worst Image | Worst Image Prompt |
|---|---|---|---|---|

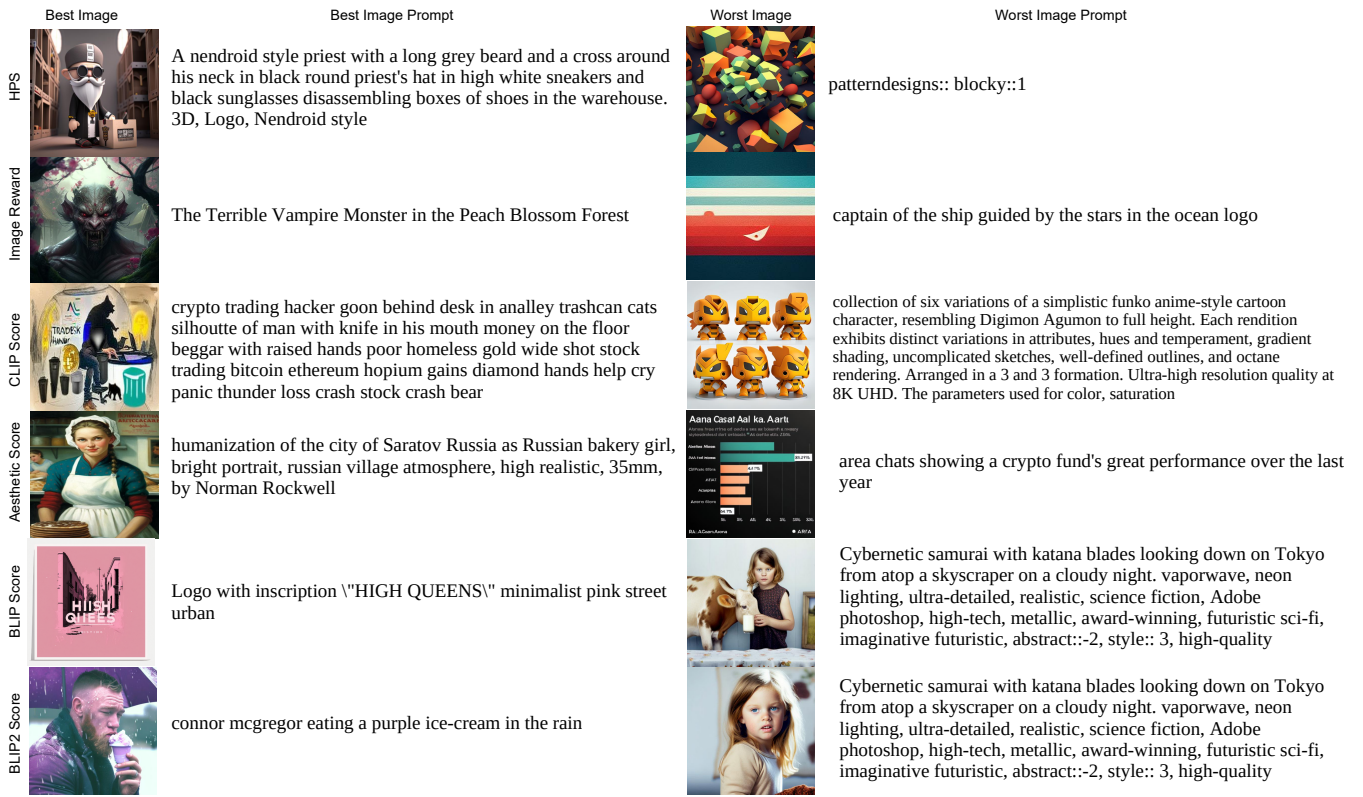

**HPS** — A nendroid style priest with a long grey beard and a cross around his neck in black round priest's hat in high white sneakers and black sunglasses disassembling boxes of shoes in the warehouse. 3D, Logo, Nendroid style / patterndesigns:: blocky::1

**Image Reward** — The Terrible Vampire Monster in the Peach Blossom Forest / captain of the ship guided by the stars in the ocean logo

**CLIP Score** — crypto trading hacker goon behind desk in analley trashcan cats silhoutte of man with knife in his mouth money on the floor beggar with raised hands poor homeless gold wide shot stock trading bitcoin ethereum hopium gains diamond hands help cry panic thunder loss crash stock crash bear / collection of six variations of a simplistic funko anime-style cartoon character, resembling Digimon Agumon to full height. Each rendition exhibits distinct variations in attributes, hues and temperament, gradient shading, uncomplicated sketches, well-defined outlines, and octane rendering. Arranged in a 3 and 3 formation. Ultra-high resolution quality at 8K UHD. The parameters used for color, saturation

**Aesthetic Score** — humanization of the city of Saratov Russia as Russian bakery girl, bright portrait, russian village atmosphere, high realistic, 35mm, by Norman Rockwell / area chats showing a crypto fund's great performance over the last year

**BLIP Score** — Logo with inscription \"HIGH QUEENS\" minimalist pink street urban / Cybernetic samurai with katana blades looking down on Tokyo from atop a skyscraper on a cloudy night. vaporwave, neon lighting, ultra-detailed, realistic, science fiction, Adobe photoshop, high-tech, metallic, award-winning, futuristic sci-fi, imaginative futuristic, abstract::-2, style:: 3, high-quality

**BLIP2 Score** — connor mcgregor eating a purple ice-cream in the rain / Cybernetic samurai with katana blades looking down on Tokyo from atop a skyscraper on a cloudy night. vaporwave, neon lighting, ultra-detailed, realistic, science fiction, Adobe photoshop, high-tech, metallic, award-winning, futuristic sci-fi, imaginative futuristic, abstract::-2, style:: 3, high-quality

Fig. 4. For each image quality score, we display the corresponding image and image prompt with the highest (left) and lowest scores (right).

| Category | Best prompt | Worst prompt |
|---|---|---|
| realistic | fat boy, on bicycle, cloud, pink dress, big arms, **realistic** style | Large Luxury Single Family neighborhood with mansions + aerial drone photo + Very Detailed + Very **Realistic** + Ultra High Definition website design, UI/UX, unreal engine, detailed, ultra high definition, 2k |
| logo | a man with top hat and smoking suit with a white tiger at his left side and an ethereum **logo** fluctuating over his right hand in a futurist cyberpunk city | a vector **logo** for a software company, infrastructures, roads, highway, vector graphics, logotype |
| photo | Goddess in gold and white ivory armor ornate with golden wings on the helmet levitating above the ground, full-body portrait + **photo**-realistic renders. ultra-detailed , 8k , dramatic epic lightning, realistic texture | Please design a cover image for my upcoming video on "Simple and Free Cash Flow Spreadsheet for Businesses". The image should feature a computer or mobile screen with the spreadsheet open, displaying the main financial data of a company. There should be a button for free download of the spreadsheet in Excel or Google Sheets in the upper right corner of the screen. At the bottom of the screen, there should be a list of the main features of the spreadsheet, such as "Daily control of inflows and outflows", "Cash movement tracking", "Expense control", and "Compatibility with mobile and desktop". The title of the video should be highlighted at the top of the screen, with the following call-to-action phrase below: "Take control of your finances with the simple and free cash flow spreadsheet for businesses". The background image can be a **photo** of a desk or a financial chart to emphasize the business aspect of the spreadsheet. Please use green and blue as the main colors to convey trust and serenity. Thank you! |
| Art | character **art**, blonde man, long blonde hair in a bun, average build, glasses, purple floral button up shirt, highly detailed, Pixar **art** style, | giant octopus 3D sculpture, in the style of cloisonn00e9 plique-00e0-jour, enamel, translucent amber and teal color **art** glass, soft dim light glows from inside octopus sculpture, intricate :: center :: full body :: wide shot :: 45 degree angle :: dark background :: solid black background |
| cartoon | create a realistic 3d **cartoon** image of a happy pug with wings and a halo of angel in paradise | a **cartoon** character about money management, cute, blue |
| anime | A cat black and white, with orange parka, and a gray cap, with a chain with the bitcoin logo, **anime** cover | Tags, The explosion of cucumber and jasmine in the iced coffee turned into Hami melon, Dreamy mountains in the style of kawase hasui, Highlight the coffee jasmine cucumber, Peter Mohrbacher, James Jean, Simon Stalenhag, and CloverWorks' style, Japanese urban pop style, Japanese 1980 vintage **anime** noise, super detail, written on Japanese paper, dynamic angles, high feeling illustrations, masterpiece, 16K, Ultra HD, best quality, perfect surreal composition, decals, explode coffee bean |
| cyberpunk | purple bear with neon **cyberpunk** lines in the face roaring, dressed with a hoodie | electric train, light snow, traffic lights, train station, **cyberpunk** city, Long shot, hyper realistic, 4K, 8k, HD, cinematic, cinematic composition:: Nikon D750::15 Halogen::250 |
| portrait | realistic **portrait** of a white goat with scientific glasses and a crown of yellow flowers on his head | stero equipment robot, unreal engine 5, Real photography, movement, realism, detailed texture, Cinematic, Color Grading, **portrait** Photography, Shot on 50mm lense, Ultra-Wide Angle, Depth of Field, hyper-detailed, beautifully color-coded, insane details, intricate details, beautifully color graded, Cinematic, Color Grading, Editorial Photography, Photography, Photoshoot, Shot on 70mm lense, Depth of Field, DOF, Tilt Blur, Shutter Speed 1/1000, F/22, White Balance, 32k, Super-Resolution, Megapixel, ProPhoto RGB, , Good, Massive, Halfrear Lighting, Backlight, Natural Lighting, Incandescent, Optical Fiber, Moody Lighting, Cinematic Lighting, Studio Lighting, Soft Lighting, Volumetric, Contre-Jour, Beautiful Lighting, Accent Lighting, Global Illumination, Screen Space Global Illumination, Ray Tracing Global Illumination, Optics, Scattering, Glowing, Shadows, Rough, Shimmering, Ray Tracing Reflections, Lumen Reflections, Screen Space Reflections, Diffraction Grading, Chromatic Aberration, GB Displacement, Scan Lines, Ray Traced, Ray Tracing Ambient Occlusion, Anti-Aliasing, FXAA, TXAA, RTX, SSAO, Shaders, OpenGL-Shaders, GLSL-Shaders, Post-Processing, Post-Production, Cel Shading, Tone Mapping, insanely detailed and intricate, hypermaximalist, elegant, realistic, super detailed, dynamic pose, photograph |
| simple | wizard in **simple** blue and white clothes with gray hair and a beard picking an apple from a tree | Simplicity: A hospital logo should be **simple** and easily recognizable. This is especially important since hospitals deal with patients who may be in distress and not be able to focus on complex designs. Color scheme: Choose a color scheme that is calming, soothing, and associated with health and wellness, such as blue, green, or white. Avoid using bright, bold colors that may be overwhelming. Fonts: Use clear and easily readable fonts, preferably sans-serif fonts, that are legible even from a distance. Images: Incorporate images that are relevant to the hospital's specialty or mission, such as a caduceus for medical institutions, or a heart for cardiac centers. Originality: Avoid copying other hospital logos, and strive to create a unique design that stands out and represents the hospital's values. With these considerations in mind, here are a few hospital logo design ideas: A stylized caduceus with a **simple** font in blue or green. An abstract symbol that represents the hospital's specialty, such as a stylized heart for a cardiac center. A logo that incorporates a hospital building with a calming color scheme. A logo that uses a **simple**, clean, and modern font with a small, stylized icon or symbol in the corner. A logo that incorporates a symbol of hope and healing, such as a dove or a lotus flower, in a calming color palette |
| illustration | Lisa Frank **illustration** of a bear smoking marijuana through a bong while sitting next to a rainbow river | The logo could feature the words Bastos Chroniclesin bold, playful font. The word Bastoscould be in a different color or font to emphasize its ironic contrast with the page's wholesome content. The letter Oin Chroniclescould be replaced with a simple **illustration** of a book or a scroll to represent storytelling. The color scheme could be bright and cheerful, such as blue and orange, to convey a sense of fun and joy. |
| landscape | An old man on top of his motorcycle, tattoos, big beard and bald, in his look a long road with a beautiful **landscape** in the background, the sun setting and the sea in the distance | ableton live :: berlin, night time, year 2030, sci-fi, cupertino, flying cars, night **sky** filled with stars, high res, 4K definition, Night, Cold Colors, Color Grading, Shot on 35mm wide angle lense, Ultra-Wide Angle, Depth of Field, hyper-detailed, beautifully color-coded, insane details, intricate details, beautifully color graded, Unreal Engine 5, Cinematic, Color Grading, Editorial Photography, Photography, Photoshoot, Landscape Shot, Depth of Field, DOF, Tilt Blur, Shutter Speed 1/1000, F/22, White Balance, 32k, Super-Resolution, Megapixel, ProPhoto RGB, VR, Lonely, Good, Massive, Halfrear Lighting, Backlight, Natural Lighting, Incandescent, Optical Fiber, Moody Lighting, Cinematic Lighting, Studio Lighting, Soft Lighting, Volumetric, Contre-Jour, Beautiful Lighting, Accent Lighting, Global Illumination, Screen Space Global Illumination, Ray Tracing Global Illumination, Optics, Scattering, Glowing, Shadows, Rough, Shimmering, Ray Tracing Reflections, Lumen Reflections, Screen Space Reflections, Diffraction Grading, Chromatic Aberration, GB Displacement, Scan Lines, Ray Traced, Ray Tracing Ambient Occlusion, Anti-Aliasing, FKAA, TXAA, RTX, SSAO, Shaders, OpenGL-Shaders, GLSL-Shaders, Post Processing, Post-Production, Cel Shading, Tone Mapping, CGI, VFX, SFX, insanely detailed and intricate, super detailed |

TABLE 2

This table shows the prompts of the worst and best image for each category

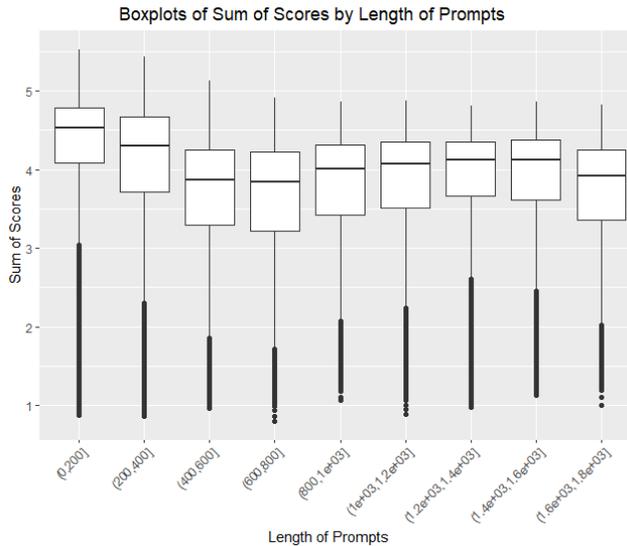

Fig. 5. This Figure shows the correlation between prompt length and the combined image quality score. For better readability we split the data based on the prompt length into nine chunks and created a boxplot for each chunk

## 2.5 Metric Correlation Analysis

To investigate the correlation among the six metrics, we applied a dimension reduction technique to the computed score vectors. This approach allows us to explore how scores vary across different categories and whether the scores of prompts within a category are similar. To this end, we reduced the six dimensions to two using t-SNE (t-distributed Stochastic Neighbor Embedding). This reduction aims to visualize high-dimensional data in a two-dimensional space, facilitating the creation of a cluster plot. By reducing the dimensionality, we hope to uncover hidden patterns and structures not immediately apparent in its original, more complex form. t-SNE excels at maintaining similarities between nearby points, making it ideal for exploring data and identifying groups of similar points. In Figure 6, we present the distribution of data points across the four largest categories: realistic, logo, photo, and art. The plots reveal that some categories, like 'realistic,' 'photo,' and 'art,' are more dispersed, forming several clusters across the space, while others, such as 'logo,' are more concentrated in specific areas. Additionally, we observe clusters representing combinations of the categories realistic, photo, and art, indicating that the prompts contain keywords from each category.

To take a closer look at these observations, we created a cluster plot shown in Figure 7 containing all four categories. The combination of categories is represented by the mixture of the categories' specific colors. The color yellow represents the combination of the category photo (with the color green) and realistic (with the color red). The results indicate that there are indeed clusters representing a category or even sub-categories, such as the orange cluster representing the combination of realistic, photo, and art, or the blue cluster representing the category logo. This suggests that the image quality metrics generally yield similar results for images within the same category.

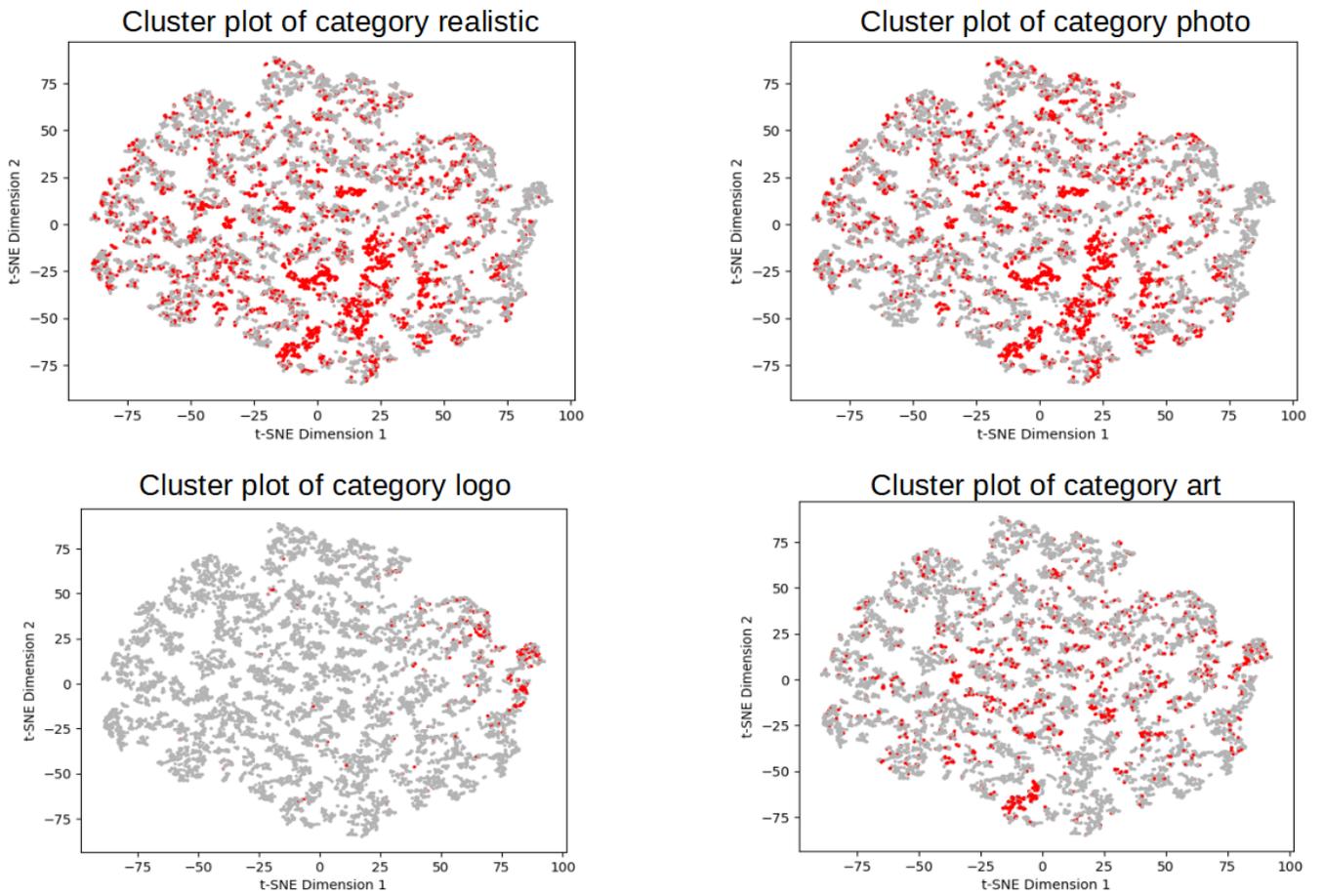

Fig. 6. These cluster plots illustrate the distribution of projected scores across four different categories. Points belonging to each category are highlighted in red, while those not belonging to any category are displayed in gray. Due to computational limitations, we considered only $10,000$ prompts for this analysis.

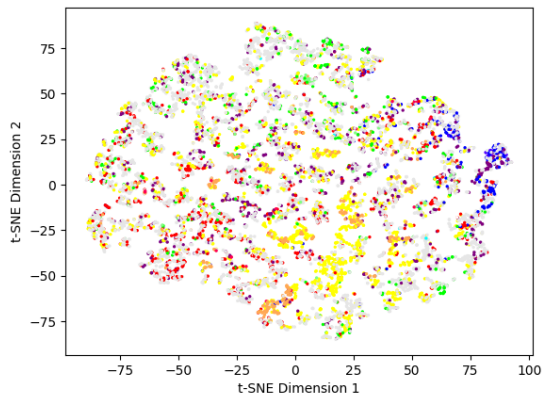

Fig. 7. This cluster plot illustrates the distribution of projected scores across four different categories. Each category is represented by a specific color: 'realistic' in red, 'logo' in blue, 'photo' in green, and 'art' in purple. The colors for combinations of these categories result from the blending of their respective colors.



\end{document}